\documentclass[11pt]{article}
\usepackage{graphicx,amsmath,amsfonts,amssymb,bm,hyperref,url,breakurl,epsfig,epsf,color,fullpage,MnSymbol,mathbbol, fmtcount,semtrans
} 

\usepackage{titlesec}

\usepackage{tikz}
\usepackage{pgfplots}
\usepackage{stackengine}

\usepackage{enumitem}

\usepackage{tikz}
\usepackage{pgfplots}
\usepackage{pgfplots}
 \usetikzlibrary{fit,shapes,positioning}

\usetikzlibrary{pgfplots.groupplots}
\usepackage{here}
\usepackage{subfig}

\usepackage{amsmath}
\usepackage{framed}
\usepackage{algorithm}
\usepackage[noend]{algpseudocode}

\titleformat{\paragraph}
{\normalfont\normalsize\bfseries}{\theparagraph}{1em}{}
\titlespacing*{\paragraph}
{0pt}{3.25ex plus 1ex minus .2ex}{1.5ex plus .2ex}

\usepackage{caption}

\usepackage[bottom,hang,flushmargin]{footmisc} 

\definecolor{darkred}{RGB}{150,0,0}
\definecolor{darkgreen}{RGB}{0,150,0}
\definecolor{darkblue}{RGB}{0,0,200}
\hypersetup{colorlinks=true, linkcolor=darkred, citecolor=darkgreen, urlcolor=black}

\newtheorem{theorem}{Theorem}
\newtheorem{lemma}{Lemma}

\newtheorem{proposition}{Proposition}
\newtheorem{definition}{Definition}

\DeclareMathOperator*{\argminA}{arg\,min}

\newcommand{\opnorm}[1]{\left\|#1\right\|}
\newcommand{\fronorm}[1]{\left\|#1\right\|_{F}}

\newcommand{\twonorm}[1]{\left\|#1\right\|_{\ell_2}}

\newcommand{\R}{\mathbb{R}}

\newcommand{\vct}[1]{\bm{#1}}
\newcommand{\mtx}[1]{\bm{#1}}
\definecolor{ejc}{RGB}{0,0,255}

\numberwithin{equation}{section} 

\def \endprf{\hfill {\vrule height6pt width6pt depth0pt}\medskip}

\title{Minimax Lower Bounds for Transfer Learning with Linear and One-hidden Layer Neural Networks} 
\date{}
\author{Seyed~Mohammadreza~Mousavi~Kalan, Zalan~Fabian, A.~Salman~Avestimehr,\\ and Mahdi~Soltanolkotabi
\\Ming Hsieh Department of Electrical Engineering, University of Southern California, CA, USA \\Email: mmousavi@usc.edu, zfabian@usc.edu, avestimehr@ee.usc.edu, soltanol@usc.edu}
\begin{document}
\maketitle
\begin{abstract}
  Transfer learning has emerged as a powerful technique for improving the performance of machine learning models on new domains where labeled training data may be scarce. In this approach a model trained for a \emph{source} task, where plenty of labeled training data is available, is used as a starting point for training a model on a related \emph{target} task with only few labeled training data. Despite recent empirical success of transfer learning approaches, the benefits and fundamental limits of transfer learning are poorly understood. In this paper we develop a statistical minimax framework to characterize the fundamental limits of transfer learning in the context of regression with linear and one-hidden layer neural network models. Specifically, we derive a lower-bound for the target generalization error achievable by any algorithm as a function of the number of labeled source and target data as well as appropriate notions of similarity between the source and target tasks. Our lower bound provides new insights into the benefits and limitations of transfer learning. We further corroborate our theoretical finding with various experiments.
  
  %Focusing on regression tasks with a linear or one-hidden layer neural network model and one-hidden layer 
 %a domain where to enable the transfer knowledge from a generalization of neural networks from 
 %Transfer learning has emerged as a way to combat the issues of deep learning models when adapting to new domains by reducing the required number of training examples. Even though many transfer learning algorithms have been developed, fundamental limits and benefits of transfer learning is still far from clear. In this paper, we develop a statistical minimax framework in the context of linear regression and one-hidden-layer neural network models to understand the limits and benefits of transfer learning. We characterize that how the transfer learning limits depend on an appropriately defined notion of similarity distance between source and target tasks and the number of source and target samples. We show that when the distance of source and target is large enough, having source samples is not helpful and in the regime that the their distance is small source samples are beneficial to reduce the generalization error in the target task. We also empirically corroborate our theoretical results via various experiments on different data sets.
\end{abstract}
\section{Introduction}
Deep learning approaches have recently enjoyed wide empirical success in many applications spanning natural language processing to object recognition. A major challenge with deep learning techniques however is that training accurate models typically requires lots of labeled data. While for many of the aforementioned tasks labeled data can be collected by using crowd-sourcing, in many other settings such data collection procedures are expensive, time consuming, or impossible due to the sensitive nature of the data. Furthermore, deep learning techniques often are brittle and do not adapt well to changes in the data or the environment. Transfer learning approaches have emerged as a way to mitigate these issues. Roughly speaking, the goal of transfer learning is to borrow knowledge from a \emph{source} domain, where lots of training data is available, to improve the learning process in a related but different \emph{target} domain. Despite recent empirical success the benefits as well as fundamental limitations of transfer learning remains unclear with many open challenges:

\emph{What is the best possible accuracy that can be obtained via any transfer learning algorithm? How does this accuracy depend on how similar the source and target domain tasks are? What is a good way to measure similarity/distance between two source and target domains? How does the transfer learning accuracy scale with the number of source and target data? How do the answers to the above questions change for different learning models?}

At the heart of answering these questions is the ability to predict the best possible accuracy achievable by any algorithm and characterize how this accuracy scales with how related the source and target data are as well as the number of labeled data in the source and target domains. In this paper we take a step towards this goal by developing statistical minimax lower bounds for transfer learning focusing on regression problems with linear and one-hidden layer neural network models. Specifically, we derive a minimax lower bound for the generalization error in the target task as a function of the number of labeled training data from source and target tasks. Our lower bound also explicitly captures the impact of the noise in the labels as well as an appropriate notion of \emph{transfer distance} between source and target tasks on the target generalization error.
%Our lower bound also depends on an appropriate notion of \emph{transfer distance} between the source and target tasks. 
Our analysis reveals that in the regime where the transfer distance between the source and target tasks is large (i.e.~the source and target are dissimilar) the best achievable accuracy mainly depends on the number of labeled training data available from the target domain and there is a limited benefit to having access to more training data from the source domain. However, when the transfer distance between the source and target domains are small (i.e.~the source and target are similar) both source and target play an important role in improving the target training accuracy. Furthermore, we provide various experiments on real data sets as well as synthetic simulations to empirically investigate the effect of the parameters appearing in our lower bound on the target generalization error.

%Our contributions are as follows: %For the regime in which source and target tasks are far from each other we show that source data is not helpful and when they are close to each other source samples can improve the generalization performance of the target task. Our contributions in this paper are summarized as follows
%\begin{itemize}
   % \item We characterize the fundamental limits achievable by \emph{any} transfer learning algorithm by %deriving a statistical lower bound for the generalization error of the target task in terms of an %appropriately defined notion of transfer distance between the source and target tasks, as well as the %number of samples available from the source and target domains.
    %\item Our analysis reveals that in the regime where the transfer distance of source and target tasks %are large (i.e.~the source and target are dissimilar) the best achievable accuracy only depends on %the number of labeled training data available from the target domain with no benefit to having access %to training data from the source domain. However, when the transfer distance between the source and %target domains are small (i.e.~the source and target are similar) both source and target play an %important role in improving the target training accuracy.
%    \item We corroborate our theoretical results by experiments on various data sets. 
%\end{itemize}

\textbf{Related work.} There is a vast theoretical literature on the problem of domain adaptation which is closely related to transfer learning \cite{blitzer2008learning,you2019towards,chen2019transferability,wu2019domain,azizzadenesheli2019regularized,shen2018wasserstein,long2016unsupervised}. The key difference is that in domain adaptation there is no labeled target data while in transfer learning a few labeled target data is available in addition to source data. Most of the existing results in the domain adaptation literature give an upper bound for the target generalization error. For instance, the papers \cite{ben2010theory,ben2007analysis} provide an upper bound on the target generalization error in classification problems in terms of quantities such as source generalization error, the optimal joint error of source and target as well as VC-dimension of the hypothesis class. A more recent work \cite{mansour2009domain} generalizes these results to a broad family of loss functions using Rademacher complexity measures. Related, \cite{zhao2019learning} derives a similar upper bound for target generalization error as in \cite{ben2010theory} but in terms of other quantities. Finally, the recent paper \cite{zhang2019bridging} generalizes the results of \cite{ben2010theory,mansour2009domain} to multiclass classification using margin loss. 

More closely related to this paper, there are a few interesting results that provide lower bounds for the target generalization error. For instance, focusing on domain adaptation the paper \cite{ben2010impossibility} provides necessary conditions for successful target learning under a variety of assumptions such as a covariate shift, similarity of unlabeled distributions, and existence of a joint optimal hypothesis. More recently, the paper \cite{hanneke2019value} defines a new discrepancy measure between source and target domains, called \textit{transfer exponent}, and  proves a minimax lower bound on the target generalization error under a relaxed covariate-shift assumption and a Bernstein class condition. \cite{zhao2019learning} derives an information theoretic lower bound on the joint optimal error of source and target domains defined in \cite{ben2010theory}. Most of the above results are based on a covariate shift assumption which requires the conditional distributions of the source and target tasks to be equal and the source and target tasks to have the same best classifier. In this paper, however, we consider a more general case in which source and target tasks are allowed to have different optimal classifiers. Furthermore, these results do not specifically study a neural network model. To the extent of our knowledge this is the first paper to develop minimax lower bounds for transfer learning with neural networks.
%explain the limitations of transfer learning in neural network models as we focus our attention on one-hidden layer neural networks and derive first minimax lower bound for the transfer learning problem in this context.%}.\MS{My comment with respect to comparison has still not been addressed}

 \section{Problem Setup}
We now formalize the transfer learning problem considered in this paper. We begin by describing the linear and one-hidden layer neural network transfer learning regression models that we study. We then discuss the minimax approach to deriving transfer learning lower bounds.

\subsection{Transfer Learning Models}\label{sec2.1}
We consider a transfer learning problem in which there are labeled training data from a source and a target task and the goal is to find a model that has good performance in the target task. Specifically, we assume we have $n_S$ labeled training data from the source domain generated according to a source domain distribution $(\vct{x}_S,\vct{y}_S)\sim \mathbb{P}$ with $\vct{x}_S\in\R^d$ representing the input/feature and $\vct{y}_S\in\R^k$ the corresponding output/label. Similarly, we assume we have $n_{T}$ training data from the target domain generated according to $(\vct{x}_T,\vct{y}_T)\sim \mathbb{Q}$ with $\vct{x}_T\in\R^d$ and $\vct{y}_T\in\R^k$. Furthermore, we assume that the features are distributed as $\vct{x}_S \sim \mathcal{N}(0,\mathbf{\Sigma_S})$, $\vct{x}_T \sim \mathcal{N}(0,\mathbf{\Sigma_T})$ with $\mtx{\Sigma}_S$ and $\mtx{\Sigma}_T \in \mathbb{R}^{d\times d}$ denoting the covariance matrices. We also assume that the labels $\vct{y}_S/\vct{y}_T$ are generated from ground truth mappings relating the features to the labels as follows
\begin{align}\label{relation}
    \vct{y}_S=f(\vct{\theta}_S;\vct{x}_S)+\vct{w}_S \ \ \ \ \ \text{and} \ \ \ \ \  \vct{y}_T=f(\vct{\theta}_T;\vct{x}_T)+\vct{w}_T
\end{align}
where $\vct{\theta}_S$ and $\vct{\theta}_T$ are the parameters of the function $f$ and $\vct{w}_S,\vct{w}_T\sim \mathcal{N}(0,\sigma^2 \mathbf{I}_{k})$ represents source/target label noise.  In this paper we focus on the following linear and one-hidden layer neural network models.

\textbf{Linear model.} In this case, we assume that $f(\vct{\theta}_S;\vct{x}_S):=f(\mtx{W}_S;\vct{x}_S)=\mtx{W}_S\vct{x}_S$ and $f(\vct{\theta}_T;\vct{x}_T):=f(\mtx{W}_T;\vct{x}_T)=\mtx{W}_T\vct{x}_T$ where $\mtx{W}_S,\mtx{W}_T \in \mathbb{R}^{k \times d}$ are two unknown matrices denoting the source/target parameters. The goal is to use the source and target training data to find a parameter matrix $\widehat{\mtx{W}}_T$ with estimated label $\widehat{\vct{y}}_T=\widehat{\mtx{W}}_T\vct{x}_T$ that achieves the smallest risk/generalization error $\mathbb{E}[\twonorm{\vct{y}_T-\widehat{\vct{y}}_T}^{2}]$.

\textbf{One-hidden layer neural network models.} We consider two different neural network models where in one the hidden-to-output layer is fixed and in the other the input-to-hidden layer is fixed. Specifically, in the first model, we assume that $f(\vct{\theta}_S;\vct{x}_S):=f(\mtx{W}_S;\vct{x}_S)=\mtx{V}\varphi(\mtx{W}_S\vct{x}_S)$ and $f(\vct{\theta}_T;\vct{x}_T):=f(\mtx{W}_T;\vct{x}_T)=\mtx{V}\varphi(\mtx{W}_T\vct{x}_T)$ where $\mtx{W}_S,\mtx{W}_T \in \mathbb{R}^{\ell \times d}$ are two unknown weight matrices, $\mtx{V}\in \mathbb{R}^{k \times \ell}$ is a fixed and known matrix, and $\varphi$ is the ReLU activation function. Similarly in the second model, we assume that $f(\vct{\theta}_S;\vct{x}_S):=f(\mtx{V}_S;\vct{x}_S)=\mtx{V}_S\varphi(\mtx{W}\vct{x}_S)$ and $f(\vct{\theta}_T;\vct{x}_T):=f(\mtx{V}_T;\vct{x}_T)=\mtx{V}_T\varphi(\mtx{W}\vct{x}_T)$ with $\mtx{V}_S,\mtx{V}_T \in \mathbb{R}^{k \times \ell}$ two unknown weight matrices and $\mtx{W}\in \mathbb{R}^{\ell \times d}$ a known matrix. In both cases the goal is to use the source and target training data to find the unknown target parameter weights ($\widehat{\mtx{W}}_T$ or $\widehat{\mtx{V}}_T$) that achieve the smallest risk/generalization error $\mathbb{E}[\twonorm{\vct{y}_T-\widehat{\vct{y}}_T}^{2}]$. Here, $\widehat{\vct{y}}_T=\mtx{V}\varphi(\widehat{\mtx{W}}_T\vct{x}_T)$ in the first model and $\widehat{\vct{y}}_T=\widehat{\mtx{V}}_T\varphi(\mtx{W}\vct{x}_T)$ in the second.

%The goal is to estimate $\mtx{W}_T$ by a matrix $\hat{\mtx{W}}_T$ by exploiting the source and target data such that the generalization error of the target task becomes small, that is, if we denote the estimated label by $\hat{\vct{y}}_T=\mtx{V}\varphi(\hat{\mtx{W}}_T\vct{x}_T)$ then the generalization error $\mathbb{E}[||\vct{y}_T-\hat{\vct{y}}_T||_F^{2}]$ is small. 
 %The goal, like the previous cases, is to come up with an estimation $\hat{\mtx{V}}_T$ such that $\mathbb{E}[||\vct{y}_T-\hat{\vct{y}}_T||_F^{2}]$ becomes small where $\hat{\vct{y}}_T=\hat{\mtx{V}}_T\varphi(\mtx{W}\vct{x}_T).$

\subsection{Minimax Framework for Transfer Learning}\label{TLM}

We now describe our minimax framework for developing lower bounds for transfer learning. As with most lower bounds, in a minimax framework we need to define a class of transfer learning problems for which the lower bound is derived. Therefore, we define $(\mathbb{P}_{\vct{\theta}_S},\mathbb{Q}_{\vct{\theta}_T})$ as a pair of joint distributions of features and labels over a source and a target task, that is, $(\vct{x}_S,\vct{y}_S)\sim \mathbb{P}_{\vct{\theta}_S}$ and $(\vct{x}_T,\vct{y}_T)\sim \mathbb{Q}_{\vct{\theta}_T}$ with the labels obeying \eqref{relation}. In this notation, each pair of a source and target task is parametrized by $\vct{\theta}_S$ and $\vct{\theta}_T$. We stress that over the different pairs of source and target tasks, $\mtx{\Sigma}_S, \mtx{\Sigma}_T,$ and $\sigma^2$ are fixed and only the parameters $\vct{\theta}_S$ and $\vct{\theta}_T$ change. %\ZF{maybe say that they are model parameters or learnable parameters?}.

As mentioned earlier, in a transfer learning problem we are interested in using both source and target training data to find an estimate $\widehat{\vct{\theta}}_T$ of $\vct{\theta}_T$ with small target generalization error. In a minimax framework, $\vct{\theta}_T$ is chosen in an adversarial way, and the goal is to find an estimate $\hat{\vct{\theta}}_T$ that achieves the smallest worst case target generalization risk $\sup \mathbb{E}_{\sim \ \text{\scriptsize source and target}\atop \text{\scriptsize samples}}\bigg[\mathbb{E}_{\mathbb{Q}_{\vct{\theta}_T}}[\twonorm{\vct{y}_T-\widehat{\vct{y}}_T}^{2}] \bigg]$. Here, the supremum is taken over the class of transfer problems under study (possible $(\mathbb{P}_{\vct{\theta}_S},\mathbb{Q}_{\vct{\theta}_T})$ pairs). We are interested in considering classes of transfer learning problems which properly reflect the difficulty of transfer learning. To this aim we need to have an appropriate notion of similarity or \emph{transfer distance} between source and target tasks. To define the appropriate measure of transfer distance we are guided by the following proposition (see Section \ref{proofprob1} for the proof) which characterizes the target generalization error for linear and one-hidden layer neural network models. 
\begin{proposition}\label{prop1}
    Let $\mathbb{Q}_{\vct{\theta}_T}$ be the data distribution over the target task with parameter $\vct{\theta}_T$ according to one of the models defined in Section \ref{TLM}. The target generalization error of an estimated model with parameter $\widehat{\vct{\theta}}_T$ is given by:
\begin{itemize}[leftmargin=*]
    \item \text{Linear model}:
    \begin{align}\label{gen}
\mathbb{E}_{\mathbb{Q}_{\vct{\theta}_T}}[\twonorm{\widehat{\vct{y}}_T-\vct{y}_T}^2]=||\mtx{\Sigma}_T^{\frac{1}{2}}(\widehat{\mtx{W}}_T-\mtx{W}_T)^T||_F^2+k\sigma^2
\end{align}
    \item One-hidden layer neural network model with fixed hidden-to-output layer:
    \begin{align}\label{gen1}
\mathbb{E}_{\mathbb{Q}_{\vct{\theta}_T}}[\twonorm{\widehat{\vct{y}}_T-\vct{y}_T}^2]\geq \frac{1}{4}\sigma^2_{\text{min}}(\mtx{V})||\mtx{\Sigma}_T^{\frac{1}{2}}(\widehat{\mtx{W}}_T-\mtx{W}_T)^T||_F^2+k\sigma^2
\end{align}
    \item One-hidden layer neural network model with fixed input-to-hidden layer:
    \begin{align}\label{gen3}
\mathbb{E}_{\mathbb{Q}_{\vct{\theta}_T}}[\twonorm{\widehat{\vct{y}}_T-\vct{y}_T}^2]=||\widetilde{\mtx{\Sigma}}_T^{\frac{1}{2}}(\widehat{\mtx{V}}_T-\mtx{V}_T)^T||_F^2+k\sigma^2
\end{align}
Here, $\widetilde{\mtx{\Sigma}}_T:=\big[\frac{1}{2}\twonorm{\vct{a}_i}\twonorm{\vct{a}_j}\frac{\sqrt{1-\gamma_{ij}^2}+(\pi-\cos^{-1}(\gamma_{ij}))\gamma_{ij}}{\pi}\big]_{ij}$ where $\vct{a}_i$ is the $i$th row of the matrix $\mtx{W}\mtx{\Sigma}_T^{\frac{1}{2}}$ and $\gamma_{ij}:=\frac{\vct{a}_i^T\vct{a}_j}{\twonorm{\vct{a}_i}\twonorm{\vct{a}_j}}$.
\end{itemize}
\end{proposition}
Proposition \ref{prop1} essentially shows how the generalization error is related to an appropriate distance between the estimated and ground truth parameters. This in turn motivates our notion of transfer distance/similarity between source and target tasks discussed next.

%Motivated by Proposition \ref{prop1}, we are now ready to formally define our notion of transfer distance/similarity between source and target tasks.
\begin{definition}\label{definition1}(Transfer distance)
 For a source and target task generated according to one of the models in Section \ref{TLM} parametrized by $\vct{\theta}_S$ and $\vct{\theta}_T$, we define the transfer distance between these two tasks as follows:
 \begin{itemize}[leftmargin=*]
     \item Linear model and one-hidden layer neural network model with fixed hidden-to-output layer:
 \begin{align}
     \rho(\vct{\theta}_S,\vct{\theta}_T)=\rho(\mtx{W}_S,\mtx{W}_T):=||\mtx{\Sigma}_T^{\frac{1}{2}}(\mtx{W}_S-\mtx{W}_T)^T||_F
 \end{align}
    \item One-hidden layer neural network model with fixed input-to-hidden layer:
    \begin{align}
     \rho(\vct{\theta}_S,\vct{\theta}_T)=\rho(\mtx{V}_S,\mtx{V}_T):=||\widetilde{\mtx{\Sigma}}_T^{\frac{1}{2}}(\mtx{V}_S-\mtx{V}_T)^T||_F
 \end{align}
 where $\widetilde{\Sigma}_T$ is defined in Proposition \ref{prop1}.
    \end{itemize}
 \end{definition}
 With the notion of transfer distance in hand we are now ready to formally define the class of pairs of distributions over source and target tasks which we focus on in this paper. 
\begin{definition}\label{class}(Class of pairs of distributions)
For a given $\Delta \in \mathbb{R}^{+}$, $\mathcal{P}_{\Delta}$ is the class of pairs of distributions over source and target tasks whose transfer distance according to Definition \ref{definition1} is less than $\Delta$. That is, $\mathcal{P}_{\Delta}=\{(\mathbb{P}_{\vct{\theta}_S},\mathbb{Q}_{\vct{\theta}_T})|\ \rho(\vct{\theta}_S,\vct{\theta}_T)\leq \Delta \}$.
\end{definition}
With these ingredients in place we are now ready to formally state the transfer learning minimax risk.
 \begin{align}\label{minmaxrisk}
\mathcal{R}_T(\mathcal{P}_{\Delta}):=\inf_{\widehat{\vct{\theta}}_T}\sup_{(\mathbb{P}_{\vct{\theta}_S}, \mathbb{Q}_{\vct{\theta}_T}) \in \mathcal{P}_{\Delta}}\mathbb{E}_{S_{\mathbb{P}_{\vct{\theta}_S}}\sim \mathbb{P}_{\vct{\theta}_S}^{1:n_{S}}}\bigg[\mathbb{E}_{S_{\mathbb{Q}_{\vct{\theta}_T}}\sim \mathbb{Q}_{\vct{\theta}_T}^{1:n_{T}}}\bigg[\mathbb{E}_{\mathbb{Q}_{\vct{\theta}_T}}[\twonorm{\vct{y}_T-\widehat{\vct{y}}_T}^{2}] \bigg]\bigg]
\end{align}
Here, $S_{\mathbb{P}_{\vct{\theta}_S}}$ and $S_{\mathbb{Q}_{\vct{\theta}_T}}$ denote i.i.d. samples $\{(\vct{x}_S^{(i)},\vct{y}_S^{(i)})\}_{i=1}^{n_S}$ and $\{(\vct{x}_T^{(i)},\vct{y}_T^{(i)})\}_{i=1}^{n_T}$ generated from the source and target distributions. We would like to emphasize that $\widehat{y}_{T}$ as defined in section \ref{relation}, is a function of samples $(S_{\mathbb{P}_{\vct{\theta}_S}},S_{\mathbb{Q}_{\vct{\theta}}})$.

\section{Main Results}\label{main-RE}
In this section, we  provide a lower bound on the transfer learning minimax risk \eqref{minmaxrisk} for the three transfer learning models defined in Section \ref{relation}. As with any other quantity related to generalization error this risk naturally depends on the size of the model and how correlated the features are in the target model. The following definition aims to capture the effective number of parameters of the model.
\begin{definition}\label{eff-dim}(Effective dimension) The effective dimension of the three models defined in Section \ref{relation} are defined as follows:
\begin{itemize}[leftmargin=*]
    \item \text{Linear model}: $D:=rank(\mtx{\Sigma}_T)k-1$,

    \item One-hidden layer neural network model with fixed hidden-to-output layer: $D:=rank(\mtx{\Sigma}_T)\ell-1$,

    \item One-hidden layer neural network model with fixed input-to-hidden layer: $D:=rank(\widetilde{\mtx{\Sigma}}_T)k-1$.
\end{itemize}
\end{definition}
Our results also depend on another quantity which we refer to as the transfer coefficient. Roughly speaking these quantities are meant to capture the relative effectiveness of a source training data from the perspective of the generalization error of the target task and vice versa.
\begin{definition}\label{def-ef-sa}(Transfer coefficients) Let $n_{S}$ and $n_{T}$ be the number of source and target training data. We define the transfer coefficients in the three models defined in Section \ref{relation} as follows 
\begin{itemize}[leftmargin=*]
    \item \text{Linear model}: $r_{S}:=\opnorm{\mtx{\Sigma}_S^{\frac{1}{2}}\mtx{\Sigma}_T^{-\frac{1}{2}}}^2$ and $r_{T}:=1$.
    \item One-hidden layer neural net with fixed output layer: $r_{S}:=\opnorm{\mtx{\Sigma}_S^{\frac{1}{2}}\mtx{\Sigma}_T^{-\frac{1}{2}}}^2 \opnorm{\mtx{V}}^2$ and $r_{T}:=\opnorm{\mtx{V}}^2$.
    \item One-hidden layer neural net model with fixed input layer: $r_{S}:=\opnorm{\widetilde{\mtx{\Sigma}}_S^{\frac{1}{2}}\widetilde{\mtx{\Sigma}}_T^{-\frac{1}{2}}}^2$ and $r_{T}:=1$. Here, $\widetilde{\mtx{\Sigma}}_S:=\big[\frac{1}{2}\twonorm{\vct{c}_i}\twonorm{\vct{c}_j}\frac{\sqrt{1-\widetilde{\gamma}_{ij}^2}+(\pi-\cos^{-1}(\widetilde{\gamma}_{ij}))\widetilde{\gamma}_{ij}}{\pi}\big]_{ij}$ where $\vct{c}_i$ is the $i$th row of $\mtx{W}\mtx{\Sigma}_S^{\frac{1}{2}}$ and $\widetilde{\gamma}_{ij}=\frac{\vct{c}_i^T\vct{c}_j}{\twonorm{\vct{c}_i}\twonorm{\vct{c}_j}}$ and $\widetilde{\mtx{\Sigma}}_T$ are defined per Proposition \ref{prop1}. 
\end{itemize}
In the above expressions $||\cdot||$ stands for the operator norm. Furthermore, we define the effective number of source and target samples as $r_{S}n_{S}$ and $r_{T}n_{T}$, respectively.
\end{definition}

With these definitions in place we now present our lower bounds  on the transfer learning minimax risk of any algorithm for the linear and one-hidden layer neural network models (see sections \ref{sketch} and \ref{appendix} for the proof).
\begin{theorem}\label{thm1}
 Consider the three transfer learning models defined in Section \ref{relation} consisting of $n_{S}$ and $n_{T}$ source and target training data generated i.i.d.~according to a class of source/target distributions with transfer distance at most $\Delta$ per Definition \ref{class}. Moreover, let $r_{S}$ and $r_{T}$ be the source and target transfer coefficients per Definition \ref{def-ef-sa}. Furthermore, assume the effective dimension $D$ per Definition \ref{eff-dim} obeys $D\geq 20$. Then, the transfer learning minimax risk \eqref{minmaxrisk} obeys the following lower bounds:
 \begin{itemize}[leftmargin=*]
    \item \text{Linear model}: $\mathcal{R}_T(\mathcal{P}_{\Delta})\geq B+k\sigma^2$.
    \item One-hidden layer neural network with fixed hidden-to-output layer: $\mathcal{R}_T(\mathcal{P}_{\Delta})\geq \frac{1}{4}\sigma^2_{\text{min}}(\mtx{V})B+k\sigma^2$.
    \item One-hidden layer neural network model with fixed input-to-hidden layer: $\mathcal{R}_T(\mathcal{P}_{\Delta})\geq B+k\sigma^2$.
\end{itemize}
Here, $\sigma_{min}(\mtx{V})$ denotes the minimum singular value of $\mtx{V}$ and
%\begin{align}
 %   \mathcal{M}_T(\mathcal{P}_{\Delta})\geq B+k\sigma^2
%\end{align}
\begin{align}\label{B}
    B:=
\begin{cases}
\frac{\sigma^2 D}{256r_{T}n_{T}},  &\text{if}\ \Delta\geq  \sqrt{\frac{\sigma^2 D\log{2}}{r_{T}n_{T}}}\\
\frac{1}{100}\Delta^2[1-0.8\frac{r_{T}n_{T}\Delta^2}{\sigma^2D}], \ \ \ \ &\text{if}\ \frac{1}{45}\sqrt{\frac{\sigma^2D}{r_{S}n_{S}+r_{T}n_{T}}}\leq \Delta < \sqrt{\frac{\sigma^2D\log{2}}{r_{T}n_{T}}}\\
\frac{\Delta^2}{1000}+\frac{6}{1000}\frac{D\sigma^2}{r_{S}n_{S}+r_{T}n_{T}}, \ \ \ \ &\text{if}\  \Delta< \frac{1}{45}\sqrt{\frac{\sigma^2D}{r_{S}n_{S}+r_{T}n_{T}}}\\
\end{cases}
\end{align}

\end{theorem}
Note that, the nature of the lower bound and final conclusions provided by the above theorem are similar for all three models. More specifically, Theorem \ref{thm1} leads to the following conclusions:
\begin{itemize}[leftmargin=*]
    \item \textbf{Large transfer distance} $(\Delta\geq \sqrt{\frac{D\sigma^2\log{2}}{r_Tn_{T}}})$. When the transfer distance between the source and target tasks is large, source samples are helpful in decreasing the target generalization error until the error reaches $\frac{\sigma^2 D}{256r_{T}n_{T}}$. Beyond this point, by increasing the number of source samples, target generalization error does not decrease further and it becomes dominated by the target samples. In other words, when the distance is large, source samples cannot compensate for target samples.
    
    \item \textbf{Moderate distance} $(\frac{1}{45}\sqrt{\frac{\sigma^2D}{r_{S}n_{S}+r_{T}n_{T}}}\leq \Delta < \sqrt{\frac{\sigma^2D\log{2}}{r_{T}n_{T}}})$. The lower bound of this regime suggests that if the distance between the source and target tasks is strictly positive, i.e $\Delta>0$, even if we have infinitely many source samples, target generalization error still does not go to zero and depends on the number of available target samples. In other words, source samples cannot compensate for the lack of target samples.  
    
    \item \textbf{Small distance} $(\Delta< \frac{1}{45}\sqrt{\frac{\sigma^2D}{r_{S}n_{S}+r_{T}n_{T}}})$. In this case, the lower bound on the target generalization error scales with $\frac{1}{r_{S}n_{S}+n_{T}r_{T}}$ where $r_{S}n_{S}$ and $r_{T}n_{T}$ are the effective number of source and target samples per Definition \ref{def-ef-sa}. Hence, when $\Delta$ is small, the target generalization error scales with the reciprocal of the total  effective number of source and target samples which means that source samples are indeed helpful in reducing the target generalization error and every source sample is roughly equivalent to $\frac{r_{S}}{r_{T}}$ target samples. Furthermore, when the distance of source and target is zero, i.e. $\Delta=0$, the lower bound reduces to $\frac{6}{1000}\frac{D\sigma^2}{r_{S}n_{S}+r_{T}n_{T}}$. Conforming with our intuition, in this case the bound resembles a non-transfer learning scenario where a combination of source and target samples are used. Indeed, the lower bound is proportional to the noise level, effective dimension and the total number of samples matching typical statistical learning lower bounds.
\end{itemize}

\section{Experiments and Numerical Results}
We demonstrate the validity of our theoretical framework through experiments on real datasets sampled from ImageNet as well as synthetic simulated data. The experiments on ImageNet data allow us to investigate the impact of transfer distance and noise parameters appearing in Theorem \ref{thm1} on the target generalization error. However, since the source and target tasks are both image classification, they are inherently correlated with each other and we cannot expect a wide range of transfer distances between them. Therefore, we carry out a more in-depth study on simulated data to investigate the effect of the number of source and target samples on the target generalization error in different transfer distance regimes. Full source code to reproduce the results can be found at \cite{imagenetcode}. %materials.%\MS{Reference missing}.
\subsection{ImageNet Experiments}
Here we verify our theoretical formulation on a subset of ImageNet, a well-known image classification dataset and show that our main theorem conforms with practical transfer learning scenarios.

\begin{figure}[t!]
  \begin{minipage}[b]{\linewidth}
    \centering
    \includegraphics[width=0.77\linewidth]{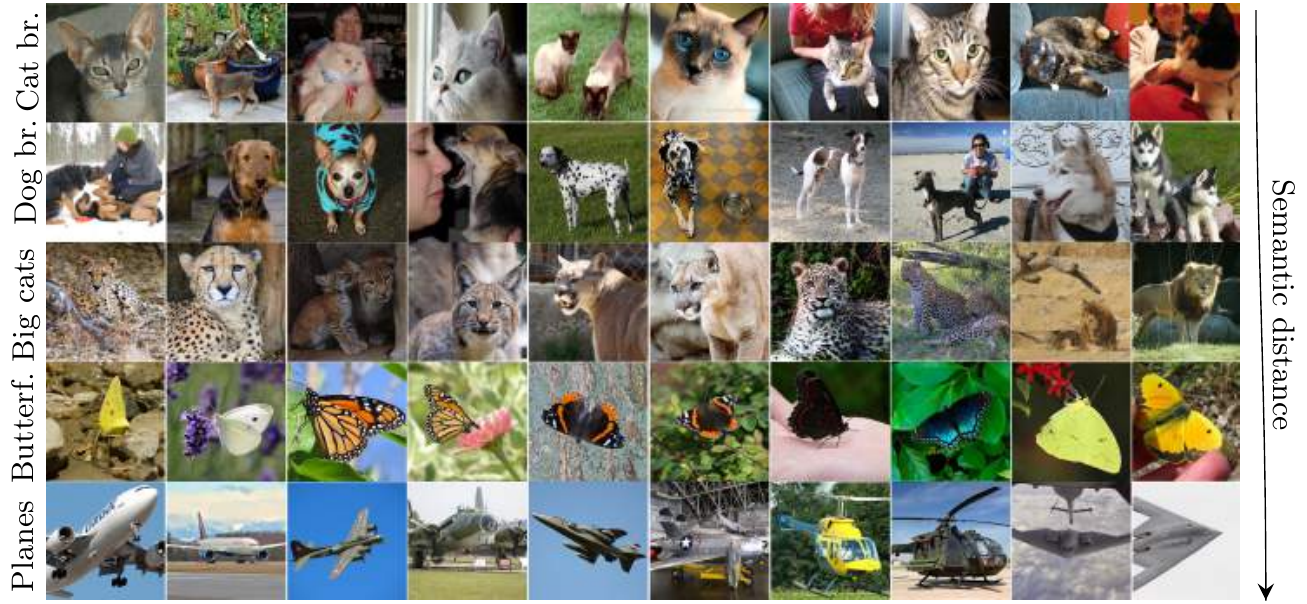}
  \end{minipage}
  \caption{Sample images from the source/target datasets derived from ImageNet. Transfer distance increases from top to bottom. \label{fig:montage}}
 \end{figure}

\textbf{Sample datasets.} We create five datasets by sub-sampling $2000$ images in five classes from ImageNet ($400$ examples per class). As depicted in Figure \ref{fig:montage}, we deliberately compile datasets covering a spectrum of semantic distances from each other in order to study the utility/effect of transfer distance on transfer learning. The picked datasets are as follows: \textit{cat breeds}, \textit{big cats}, \textit{dog breeds}, \textit{butterflies}, \textit{planes}. For details of the classes in each dataset please refer to the code provided in \cite{imagenetcode}. We pass the images through a VGG16 network pretrained on ImageNet with the fully connected top classifier removed and use the extracted features instead of the actual images. We set aside $10 \%$ of the dataset as test set. Furthermore, $10 \%$ of the remaining data is used for validation and $90 \%$ for training. In the following we fix identifying \textit{cat breeds} as the source task and the four other datasets as target tasks.

\begin{table}[t!]
\centering
\begin{tabular}{ |p{4.5cm}||p{3cm}|p{2.5cm}|p{2.5cm}|  }
 \hline
 Source / target task& $\rho(source, target)$ &Validation loss & Noise level ($\sigma$)\\
 \hline
 \textit{cat breeds} / \textit{dog breeds}   & 11.62    &0.2194 & 0.2095 \\
 \textit{cat breeds} / \textit{big cats}&   12.35  & 0.1682 & 0.1834\\
 \textit{cat breeds} / \textit{butterflies} &13.48 & 0.1367 & 0.1653\\
 \textit{cat breeds} / \textit{planes}    &16.41 & 0.1450 &0.1703\\
 \hline 
\end{tabular}
\caption{Transfer distance and noise level for various source-target pairs.\label{numeric:table1}}
\end{table}
\begin{figure}
  \begin{minipage}[b]{0.45\linewidth}
    \centering
    \includegraphics[width=0.97\linewidth]{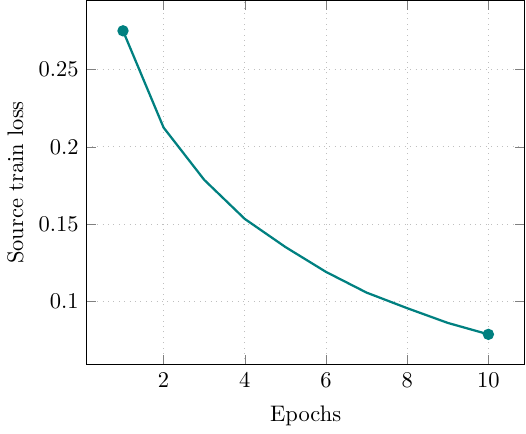}
  \end{minipage}
%    \begin{minipage}[b]{0.49\linewidth}
%    \centering
%    \includegraphics[width=0.95\linewidth]{plots/train_acc.pdf}
%  \end{minipage}
\begin{minipage}[b]{0.45\linewidth}
    \centering
    \includegraphics[width=\linewidth]{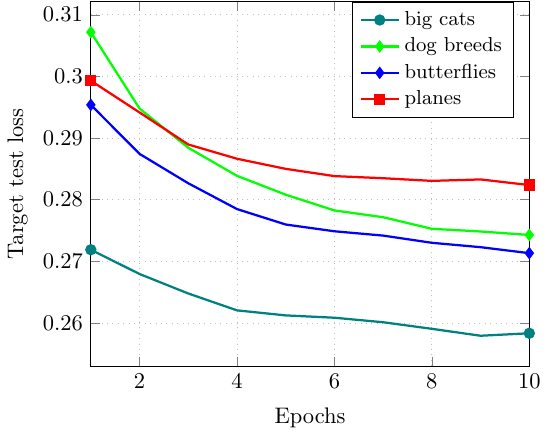}
 \end{minipage}
%    \begin{minipage}[b]{0.49\linewidth}
%    \centering
%    \includegraphics[width=\linewidth]{plots/val_acc.pdf}
%  \end{minipage}
\caption{Train and test loss of a one-hidden layer network trained on \textit{cat breeds} dataset.}
\label{fig:exp2}
\end{figure}

\textbf{Training.} We trained a one-hidden layer neural network for each dataset. To facilitate fair comparison between the trained models in weight-space, we fixed a random initialization of the hidden-to-output layer shared between all networks and we only trained over the input-to-hidden layer (in accordance with the theoretical formulation). Moreover, we used the same initialization of input-to-hidden weights. We trained a separate model on each of the five datasets on MSE loss with one-hot encoded labels. We use an Adam optimizer with a learning rate of $0.0001$ and train for $100$ epochs or until the network reaches $99.5 \%$ accuracy whichever occurs first. The target noise levels are calculated based on the average loss of the trained ground truth models on the target validation set (note that this average loss equals $k\sigma^2=5\sigma^2$).

\textbf{Results.} First, we calculate the transfer distance from Definition \ref{definition1} between the model trained on the source task (\textit{cat breeds}) and the other four models trained on target tasks by fitting a ground truth model to each task using complete training data. Our results depicted in Table \ref{numeric:table1} demonstrate that the introduced metric strongly correlates with perceived semantic distance. The closest tasks, \textit{cat breeds} and \textit{dog breeds}, are both characterized by pets with similar features, and with humans frequently appearing in the images.  Images in the second closest pair, \textit{cat breeds} and \textit{big cats}, include animals with similar features, but \textit{big cats} have more natural scenes and less humans compared with \textit{dog breeds}, resulting in slightly higher distance from the source task. As expected, \textit{cat breeds}- \textit{butterflies} distance is significantly higher than in case of the previous two targets, but they share some characteristics such as the presence of natural backgrounds. The largest distance is between \textit{cat breeds} and \textit{planes}, which is clearly the furthest task semantically as well.  

Our next set of experiments focuses on checking whether the transfer distance is indicative of transfer target risk/generalization error. To this aim we use a very simple transfer learning approach where we use only source data to train a one-hidden layer network as described before and measure its performance on the target tasks. Note that the network has never seen examples of the target dataset. Figure \ref{fig:exp2} depicts how train and test loss evolved over the training process. We stop after $10$ epochs when validation losses on target tasks have more or less stabilized. The results closely match our expectations from Theorem \ref{thm1}. Based on Table \ref{numeric:table1} the noise level of ground truth models for \textit{big cats}, \textit{butterflies} and \textit{planes} are about the same and therefore their test loss follows the same ordering as their distances from the source task (see Table \ref{numeric:table1}). Moreover, even though \textit{dog breeds} has the lowest distance from the source task, it is also the noisiest. The lower bound in Theorem \ref{thm1} includes an additive noise term, and therefore the change in ordering between \textit{dog breeds} and \textit{butterflies} is justified by our theory and demonstrates the effect of the target task noise level on generalization.
\subsection{Numerical Results}\label{numresults}
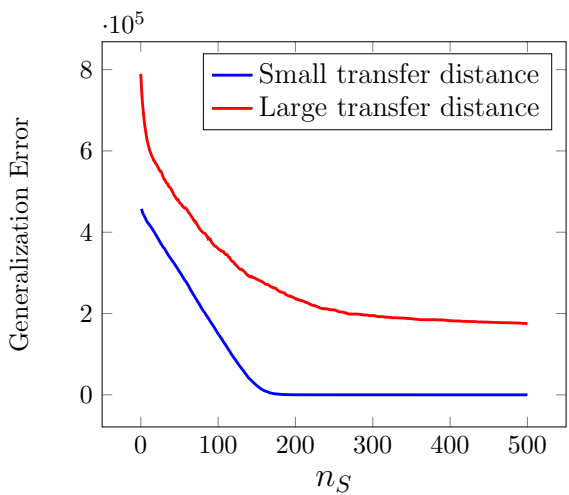
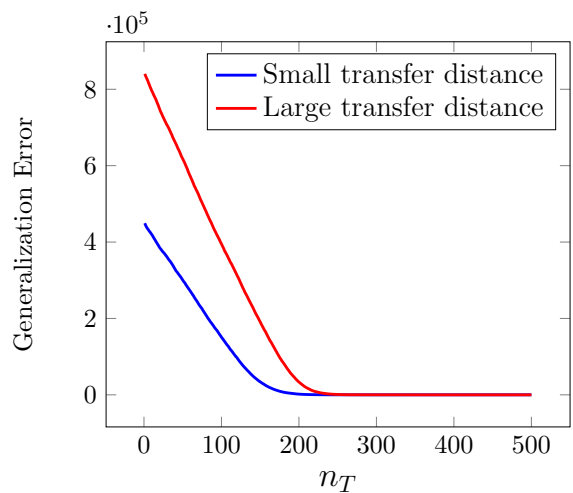
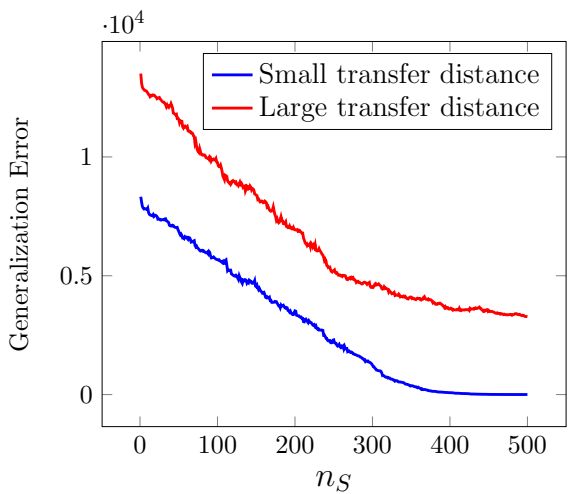
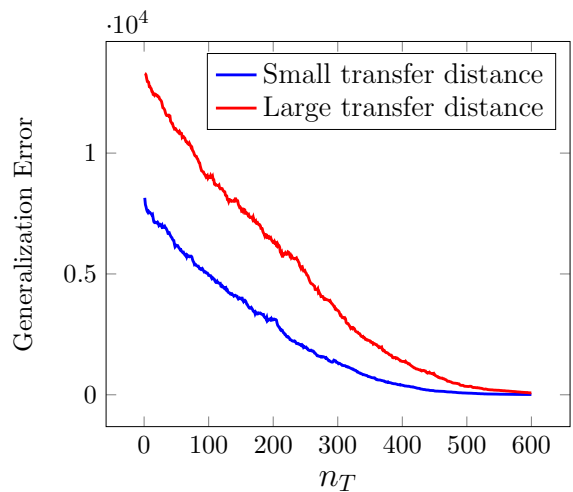
\begin{figure}
\centering
\subfloat[Linear model with $n_T=50$]{
 \begin{tikzpicture}[scale=.9]
\begin{axis}[
    title={},
    xlabel={\Large$n_{S}$},
    ylabel={Generalization Error},
    legend pos=north east,
]
\addplot[line width=1.2pt,
    color=blue,
    ]
    table {l_np_small_averagerun.dat};
    \addlegendentry{\large\text{Small transfer distance}}
    \addplot[line width=1.2pt,
    color=red,
    ]
    table {L_np_large.dat};
    \addlegendentry{\large\text{Large transfer distance}}
\end{axis}
\end{tikzpicture}}
\hspace{10mm}
\subfloat[Linear model with $n_S=50$]{
 \begin{tikzpicture}[scale=.9]
\begin{axis}[
    title={},
    xlabel={\Large$n_{T}$},
    ylabel={Generalization Error},
    legend pos=north east,
]
\addplot[line width=1.2pt,
    color=blue,
    ]
    table {l_nq_small_averagerun.dat};
    \addlegendentry{\large\text{Small transfer distance}}
    \addplot[line width=1.2pt,
    color=red,
    ]
    table {l_nq_large_averagerun.dat};
    \addlegendentry{\large\text{Large transfer distance}}
\end{axis}

\end{tikzpicture}}

\subfloat[Neural network model with $n_{T}=50$]{
 \begin{tikzpicture}[scale=.9]
 \begin{axis}[
    title={},
    xlabel={\Large$n_{S}$},
    ylabel={Generalization Error},
    legend pos=north east,
]
\addplot[line width=1.2pt,
    color=blue,
    ]
    table {nl_np_small_averagerun.dat};
    \addlegendentry{\large\text{Small transfer distance}}
    \addplot[line width=1.2pt,
    color=red,
    ]
    table {nl_np_large_averagerun.dat};
    \addlegendentry{\large\text{Large transfer distance}}
\end{axis}

\end{tikzpicture}}
\hspace{10mm}
\subfloat[Neural network model with $n_{S}=50$]{
 \begin{tikzpicture}[scale=.9]
\begin{axis}[
    title={},
    xlabel={\Large$n_{T}$},
    ylabel={Generalization Error},
    legend pos=north east,
]
\addplot[line width=1.2pt,
    color=blue,
    ]
    table {NL_nq_small_averagerun.dat};
    \addlegendentry{\large\text{Small transfer distance}}
    \addplot[line width=1.2pt,
    color=red,
    ]
    table {NL_nq_large_averagerun.dat};
    \addlegendentry{\large\text{Large transfer distance}}
\end{axis}
\end{tikzpicture}}
\caption{Target generalization error for a linear model ((a) and (b)) and a neural network model with fixed hidden-to-output layer ((c) and (d)).}
\label{Fig2}
\end{figure}
In this section we perform synthetic numerical simulations in order to carefully cover all regimes of transfer distance from our main theorem, and show how the target generalization error depends on the number of source and target samples in different regimes.

\textbf{Experimental setup 1.} First, we generate data according to the linear model with parameters $d=200,  k=30, \sigma=1, \mtx{\Sigma}_S=2\cdot I_{d}, \mtx{\Sigma}_T=I_{d}$. Then we generate the source parameter matrix $\mtx{W}_S\in \mathbb{R}^{k\times d}$ with elements sampled from $\mathcal{N}(0,10)$. Furthermore, we generate two target parameter matrices $\mtx{W}_{T_1}$ and $\mtx{W}_{T_2} \in \mathbb{R}^{k\times d}$ for tasks $T_1$ and $T_2$ such that $\mtx{W}_{T_1}=\mtx{W}_S+\mtx{M}_1$ and $\mtx{W}_{T_2}=\mtx{W}_S+\mtx{M}_2$ where the elements of $\mtx{M}_1, \mtx{M}_2$ are sampled from $\mathcal{N}(0,10^{-3})$ and $\mathcal{N}(0,3.6\times 10^{5})$ respectively.
 Similarly for the one-hidden layer neural network model when the the output layer is fixed, we set the parameters $k=1, \ell=30, d=200, \sigma=1, \mtx{\Sigma}_S=2\cdot I_{d}, \mtx{\Sigma}_T=I_{d}$ and $V=\mathbf{1}_{k\times \ell}$. We also use the same $\mtx{W}_{S}, \mtx{W}_{T_1}, \mtx{W}_{T_2}$ as in the linear model. We note that the transfer distance between the source task to target task $T_1$ is small but the transfer distance between the source task to target task $T_2$ is large ($\rho(\mtx{W}_S,\mtx{W}_{T_1})=.0183$ and $\rho(\mtx{W}_S,\mtx{W}_{T_2})=116.694$). %\ZF{Do we have to include all the exact numerical values/dimensions of these parameters here? It might be a bit confusing/distracting to the reader and we might as well just have the exact numbers in the appendix?}
 
 \textbf{Training approach 1.} We test the performance of a simple transfer learning approach. Given $n_{S}$ source samples and $n_{T}$ target samples, we estimate $\widehat{\mtx{W}}_T$ by minimizing the weighted empirical risk 
 \begin{align}\label{weight}
 \min_{\mtx{W}}\text{ }\frac{1}{2n_T}\sum_{i=1}^{n_T}\twonorm{f(\mtx{W};\vct{x}_T^{(i)})-\vct{y}_T^{(i)}}^2+\frac{\lambda}{2n_S}\sum_{j=1}^{n_S}\twonorm{f(\mtx{W};\vct{x}_S^{(j)})-\vct{y}_S^{(j)}}^2
 \end{align}
We then evaluate the generalization error by testing the estimated model $\widehat{\mtx{W}}_T$ on 200 unseen test data points generated by the target model. All reported plots are the average of $10$ trials.

 \textbf{Results 1.} Figure \ref{Fig2} (a) depicts the target generalization error for target tasks $T_1$ and $T_2$ for the linear model for different $n_{S}$ values with $\lambda=1$ and $n_T=50$. Figure \ref{Fig2} (b) depicts the target generalization error for tasks $T_1$ and target $T_2$ for the linear model for different $n_{T}$ values with the number of source samples fixed at $n_S=50$. Here, we set $\lambda=1$ for target task $T_1$, where the transfer distance from source is small, and $\lambda=.001$ for target task $T_2$, where the transfer distance from source is large.  Figures \ref{Fig2} (c) and \ref{Fig2} (d) have the same settings as in Figures \ref{Fig2} (a) and \ref{Fig2} (b) but we use a one-hidden layer neural network model with fixed hidden-to-output weights in lieu of the linear model.

 Figures \ref{Fig2} (a) and (c) clearly demonstrate that when the transfer distance between the source and target tasks is large, increasing the number of source samples is not helpful beyond a certain point. In particular, the target generalization error starts to saturate and does not decrease further. Stated differently, in this case the source samples cannot compensate for the target samples. This observation conforms with our main theoretical result. Indeed, when the transfer distance $\Delta$ is large, $B$ is lower bounded by $\frac{\sigma^2 D}{256r_{T}n_{T}}$ which is independent of the number of source samples $n_S$. Furthermore, these figures also demonstrate that when the transfer distance is small, increasing the number of source samples is helpful and results in lower target generalization error. This also matches our theoretical results as when the transfer distance $\Delta$ is small, the target generalization error is proportional to $\frac{D\sigma^2}{r_{S}n_{S}+r_{T}n_{T}}$.
 
Figures \ref{Fig2} (b) and (d) indicate that regardless of the transfer distance between the source and target tasks the target generalization error steadily decreases as the number of target samples increases. This is a good match with our theoretical results as  $n_{T}$ appears in the denominator of our lower bound in all three regimes. %Furthermore, the supplementary contains an additional simulation where we depict the target generalization error as a function of the transfer distance of source and target.

To further investigate the effect of transfer distance between the source and target on the target generalization error we consider another set of experiments below.

\textbf{Experimental setup 2.} For the linear model, we use the parameters $d=50,  k=30, \sigma=0.3,  \mtx{\Sigma}_S=2\cdot I_d$, and $\mtx{\Sigma}_T=I_d$. We generate the target parameter $\mtx{W}_T \in \mathbb{R}^{k \times d}$ with entries generated i.i.d.~$\mathcal{N}(0,10)$. To create different transfer distances between the source and target data we then generate the source parameter $\mtx{W}_S\in \mathbb{R}^{k \times d}$ as $\mtx{W}_S=\mtx{W}_T+i\cdot\mtx{M}$ where the elements of the matrix $\mtx{M}$ are sampled from $\mathcal{N}(0,10^{-4})$ and $i$ varies between $1$ and $140000$ in increments of $400$. Similarly for the one-hidden layer neural network model when the the output layer is fixed, we pick parameter values $k=1, \ell=30, d=50, \sigma=0.3, \mtx{\Sigma}_S=2\cdot \mtx{I}_{d}$,  and $\mtx{\Sigma}_T=\mtx{I}_{d}$ and set all of the entries of $V$ equal to one. Furthermore, we use the same source and target parameters $\mtx{W}_S$ and $\mtx{W}_T$ as in the linear model.

\textbf{Training approach 2.} Given $n_S=300$ and $n_T=20$ source and target samples we minimize the weighted empirical risk \eqref{weight}. In this experiment we pick $\lambda\in\{0,\frac{1}{4},\frac{1}{2},\frac{3}{4},1\}$ that minimizes a validation set consisting of $50$ data points created from the same distribution as the target task. Finally we test the estimated model on $200$ unseen target test data points. The reported numbers are based on an average of $20$ trials    .

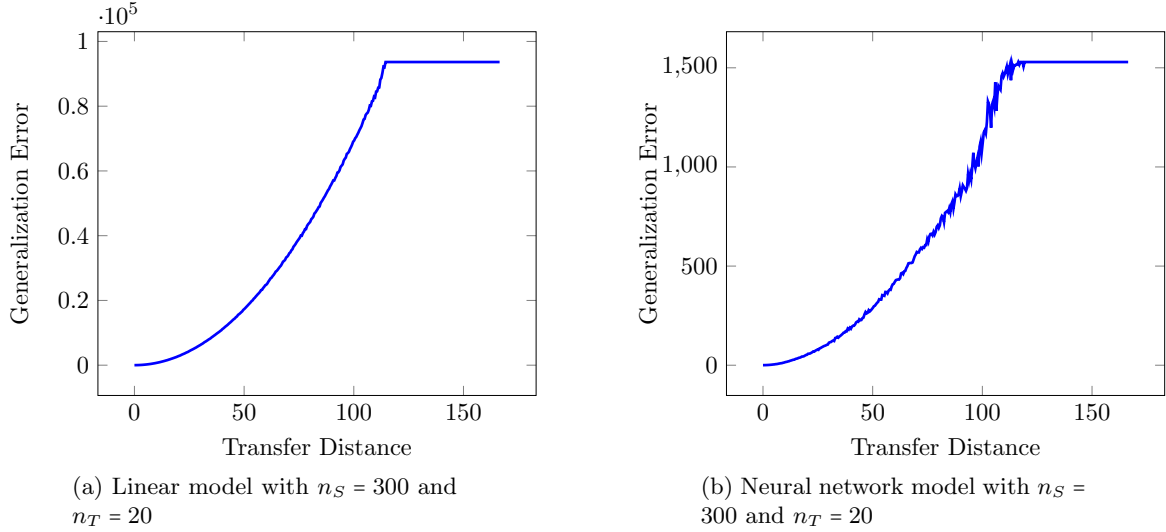
\begin{figure}
\centering
\subfloat[Linear model with $n_S=300$ and $n_T=20$]{
 \begin{tikzpicture}[scale=.85]
\begin{axis}[
    title={},
    xlabel={\text{Transfer Distance}},
    ylabel={Generalization Error},
    legend pos=north east,
]
\addplot[line width=1.2pt,
    color=blue,
    ]
    table {lnew_delta_averagerun_cross_1lam.dat};
    %\addlegendentry{\large\text{Small transfer distance}}
\end{axis}
\end{tikzpicture}}
\hspace{10mm}
\subfloat[Neural network model with $n_S=300$ and $n_T=20$]{
 \begin{tikzpicture}[scale=.85]
\begin{axis}[
    title={},
    xlabel={\text{Transfer Distance}},
    ylabel={Generalization Error},
    legend pos=north east,
]
\addplot[line width=1.2pt,
    color=blue,
    ]
    table {nlnew_delta_averagerun_cross_1lam.dat};
    %\addlegendentry{\large\text{Small transfer distance}}
\end{axis}

\end{tikzpicture}}
\caption{Target generalization error for a linear model (a) and a neural network model with fixed hidden-to-output layer (b).}
\label{Fig-delta}
\end{figure}

\textbf{Results 2.} Figure \ref{Fig-delta} depicts the target generalization error as a function of the transfer distance between the source and target in the linear and neural network models. This figure clearly shows that when the transfer distance is small, the generalization error has a quadratic growth. However, as the distance increases the error saturates which matches the behavior of $\Delta$ predicted by our lower bounds.

\newpage
\section{Proof outline and proof of Theorem \ref{thm1} in the linear model}\label{sketch}
In this section we present a sketch of the proof of Theorem \ref{thm1} for the linear model. The proof for the neural network models follow a similar approach and appear in Sections \ref{proofmodel2} and \ref{proofmodel3}.

Note that by Proposition \ref{prop1}, the generalization error is given by
\begin{align*}
\mathbb{E}_{\mathbb{Q}_{\vct{\theta}_T}}[\twonorm{\widehat{\vct{y}}_T-\vct{y}_T}^2]=||\mtx{\Sigma}_T^{\frac{1}{2}}(\widehat{\mtx{W}}_T-\mtx{W}_T)^T||_F^2+k\sigma^2.
\end{align*}
Therefore, in order to find a minimax lower bound on the target generalization error, it suffices to find a lower bound for the following quantity
\begin{align}\label{minm}
\mathcal{R}_T(\mathcal{P}_{\Delta}&;\phi \circ \rho)\nonumber\\
&:=\inf_{\widehat{\mtx{W}}_T}\sup_{(\mathbb{P}_{\mtx{W}_S}, \mathbb{Q}_{\mtx{W}_T}) \in \mathcal{P}_{\Delta}}\mathbb{E}_{S_{\mathbb{P}_{\mtx{W}_S}}\sim \mathbb{P}_{\mtx{W}_S}^{1:n_{\mathbb{P}}}}\big[\mathbb{E}_{S_{\mathbb{Q}_{\mtx{W}_T}}\sim \mathbb{Q}_{\mtx{W}_T}^{1:n_{\mathbb{Q}}}}\big[\phi(\rho(\widehat{\mtx{W}}_T(S_{\mathbb{P}_{\mtx{W}_S}},S_{\mathbb{Q}_{\mtx{W_T}}}),\mtx{W_T}))\big]\big]
\end{align}
where $\phi(x)=x^2$ for $x\in \mathbb{R}$ and $\rho$ is per Definition \ref{definition1}. 
By using well-known techniques from the statistical minimax literature we reduce the problem of finding a lower bound to a hypothesis testing problem (e.g.~see \cite[Chapter 15]{wainwright2019high}). Since we are estimating the target parameter, i.e. $\mtx{W}_T$, to apply this framework we need to pick $N$ pairs of distributions from the set $\mathcal{P}_{\Delta}$ such that their target parameters are $2\delta$-separated by the transfer distance per Definition \ref{definition1}. To be more precise, we pick $N$ arbitrary pairs of distributions from $\mathcal{P}_{\Delta}$ : $$(\mathbb{P}_{\mtx{W}_S^{(1)}},\mathbb{Q}_{\mtx{W}_T^{(1)}}),...,(\mathbb{P}_{\mtx{W}_S^{(N)}},\mathbb{Q}_{\mtx{W}_T^{(N)}})$$ such that
\begin{align*}
    \rho(\mtx{W}_T^{(i)},\mtx{W}_T^{(j)})\geq 2\delta \ \  \text{for each}\ i\neq j \in [N]\times [N] \ \ (2\delta\text{-separated set})
\end{align*}
and 
\begin{align*}
    \rho(\mtx{W}_S^{(i)},\mtx{W}_T^{(i)})\leq \Delta \ \  \text{for each}\ i\in [N] \ \ (\text{as they belong to} \ \mathcal{P}_{\Delta})
\end{align*}
With these $N$ distribution pairs in place we can follow a proof similar to that of \cite[Proposition 15.1]{wainwright2019high} to reduce the minimax problem to a hypothesis test problem. In particular, consider the following $N$-array hypothesis testing problem:% by $\{(\mathbb{P}_{\mtx{W}_S^{(i)}},\mathbb{Q}_{\mtx{W}_T^{(i)}}), i=1,2,...,N\}$:
\begin{itemize}
    \item $J$ is the uniform distribution over the index set $[N]:=\{1,2,...,N\}$
    \item Given $J=i$, generate $n_S$ i.i.d. samples from $\mathbb{P}_{\mtx{W}_S^{(i)}}$ and $n_T$ i.i.d. samples from $\mathbb{Q}_{\mtx{W}_T^{(i)}}$. 
\end{itemize}
Here the goal is to find the true index using $n_S+n_T$ available samples by a testing function $\psi$ from the samples to the indices.  

Let $E$ and $F$ be random variables such that $E | \{J=i\}\sim \mathbb{P}_{\mtx{W}_S^{(i)}}$ and $F | \{J=i\}\sim \mathbb{Q}_{\mtx{W}_T^{(i)}}$. Furthermore, let $Z_{\mathbb{P}}$ and $Z_{\mathbb{Q}}$ consist of $n_{S}$ independent copies of random variable $E$ and $n_{T}$ independent copies of random variable $F$, respectively. In this setting, by slightly modifying the \cite[Proposition 15.1]{wainwright2019high} we can conclude that 
$$\mathcal{R}_T(\mathcal{P}_{\Delta};\phi \circ \rho)\geq \phi(\delta)\frac{1}{N}\sum_{i=1}^{N}\text{Prob}(\psi(Z_{\mathbb{P}},Z_{\mathbb{Q}})\neq i)$$
where $$\psi(Z_{\mathbb{P}},Z_{\mathbb{Q}}):=\argminA_{n \in [N]}\rho(\widehat{\mtx{W}}_T,\mtx{W}_T^{n}).$$ Furthermore, by using Fano's inequality we can conclude that 
\begin{align}\label{minmax1}
\mathcal{R}_T(\mathcal{P}_{\Delta};\phi \circ \rho)&\geq \phi(\delta)\frac{1}{N}\sum_{i=1}^{N}\mathbb{P}\Big\{\psi(Z_{\mathbb{P}},Z_{\mathbb{Q}})\neq i\Big\}\nonumber\\
&\geq \phi(\delta)\left(1-\frac{I(J;(Z_{\mathbb{P}},Z_{\mathbb{Q}}))+\log{2}}{\log{N}}\right)\nonumber\\
&\geq \phi(\delta)\left(1-\frac{I(J;Z_{\mathbb{P}})+I(J;Z_{\mathbb{Q}})+\log{2}}{\log{N}}\right)\nonumber\\
&\geq \phi(\delta) \left(1-\frac{n_{S}I(J;E)+n_{T}I(J;F)+\log{2}}{\log{N}}\right).
\end{align}
 Here the third inequality is due to the fact that given $J=i$, $Z_{\mathbb{P}}$ and $Z_{\mathbb{Q}}$ are independent. To continue further, note that we can bound the mutual information by the following KL-divergences
\begin{align}\label{info}
    I(J;E)&\leq \frac{1}{N^2}\sum_{i,j}D_{KL}(\mathbb{P}_{\mtx{W}_S^{(i)}}||\mathbb{P}_{\mtx{W}_S^{(j)}})\nonumber\\
    I(J;F)&\leq \frac{1}{N^2}\sum_{i,t}D_{KL}(\mathbb{Q}_{\mtx{W}_T^{(i)}}||\mathbb{Q}_{\mtx{W}_T^{(j)}}).
\end{align}
In the next lemma, proven in Section \ref{prooflemma1},  we explicitly calculate the above KL-divergences. 
\begin{lemma}\label{KLdiv}
Suppose that $\mathbb{P}_{\mtx{W}_S^{(i)}}$ and $\mathbb{P}_{\mtx{W}_S^{(j)}}$ are the joint distributions of features and labels in a source task and $\mathbb{Q}_{\mtx{W}_T^{(i)}}$ and $\mathbb{Q}_{\mtx{W}_T^{(j)}}$ are joint distributions of features and labels in a target task as defined in Section \ref{sec2.1} for the linear model. Then  $D_{KL}(\mathbb{P}_{\mtx{W}_S^{(i)}}||\mathbb{P}_{\mtx{W}_S^{(j)}})=\frac{||\mtx{\Sigma}_S^{\frac{1}{2}}(\mtx{W}_S^{(i)}-\mtx{W}_S^{(j)})^T||_F^2}{2\sigma^2}$ and $D_{KL}(\mathbb{Q}_{\mtx{W}_T^{(i)}}||\mathbb{Q}_{\mtx{W}_T^{(j)}})=\frac{||\mtx{\Sigma}_T^{\frac{1}{2}}(\mtx{W}_T^{(i)}-\mtx{W}_T^{(j)})^T||_F^2}{2\sigma^2}.$
\end{lemma}
In the following two lemmas, we use local packing techniques to further simplify \eqref{minmax1} using \eqref{info} and find minimax lower bounds in different transfer distance regimes. We defer the proof of these lemmas to Sections \ref{prooflemma2} and \ref{prooflemma3}.
\begin{lemma}\label{b1}
    Assume $\Delta\geq \sqrt{\frac{\sigma^2D\log{2}}{r_{T}n_{T}}}$, where $n_T$ is the number of target samples and $D$ and $r_T$ are defined per Definitions \ref{eff-dim} and \ref{def-ef-sa}. Then we have the following lowerbound
\begin{align}
   \mathcal{R}_T(\mathcal{P}_{\Delta};\phi \circ \rho)\geq \frac{\sigma^2D}{256r_Tn_{T}}.
\end{align}
Furthermore, if $\Delta<\sqrt{\frac{\sigma^2D\log{2}}{r_{T}n_T}}$ then
\begin{align}
   \mathcal{R}_T(\mathcal{P}_{\Delta};\phi \circ \rho)\geq \frac{1}{100}\Delta^2\left(1-0.8\frac{r_Tn_{T}\Delta^2}{\sigma^2D}\right).
\end{align} 
\end{lemma}
\begin{lemma}\label{b2}
Assume we have access to $n_{S}$ source samples as well as $n_{T}$ target samples and the transfer distance obeys $\Delta\leq \frac{1}{45}\sqrt{\frac{\sigma^2D}{r_{S}n_{S}+r_{T}n_{T}}}$, where $D$, $r_S$, and $r_T$ are per Definitions \ref{eff-dim} and \ref{def-ef-sa}. Then,
\begin{align}
   \mathcal{R}_T(\mathcal{P}_{\Delta};\phi \circ \rho)\geq \frac{\Delta^2}{1000}+\frac{6}{1000}\frac{D\sigma^2}{r_{S}n_{S}+r_{T}n_{T}}.
\end{align}
\end{lemma}
The proof of the lower bound in Theorem \ref{thm1} is complete by combining Lemmas \ref{b1} and \ref{b2}.

\section{Acknowledgements}
This material is based upon work supported by Defense Advanced Research Projects Agency (DARPA) under the Learning with Less Labels (LwLL) program. This work is also partially supported by NSF grants CCF-1703575, CIF-1846369 and CIF-1813877, ONR Award N00014-16-1-2189, ARO award W911NF1810400, AFOSR award FA9550-18-1-0078, a Packard Fellowship in Science and Engineering, a Sloan Research Fellowship in Mathematics, and a Google faculty research award. The views, opinions, and/or findings expressed are those of the author(s) and should not be interpreted as representing the official views or policies of the Department of Defense or the U.S. Government.
\bibliographystyle{IEEEtran}

\bibliography{iclr2019_conference}
\newpage
\section{Appendix}\label{appendix}
\subsection{Calculating the Generalization Errors ( Proof of Proposition \ref{prop1})}\label{proofprob1}
\begin{itemize}[leftmargin=*]
    \item \textbf{Linear model:}
\end{itemize}
By expanding the expression we get
\begin{align}\label{linear-gen}
    \mathbb{E}_{\mathbb{Q}_{\mtx{\theta}_T}}[\twonorm{\widehat{\vct{y}}_T-\vct{y}_T}^2]&=\mathbb{E}[\twonorm{\widehat{\vct{W}}_T\vct{x}_T-\mtx{W}_T\vct{x}_T-w_T}^2]\nonumber\\
    &=\mathbb{E}[\twonorm{\widehat{\mtx{W}}_T\vct{x}_T-\mtx{W}_T\vct{x}_T}^2]+k\sigma^2\nonumber\\
    &=\mathbb{E}[\vct{x}_T^T(\mtx{W}_T-\widehat{\mtx{W}_T})^T(\mtx{W}_T-\widehat{\mtx{W}_T})\vct{x}_T]+k\sigma^2\nonumber\\
    &=\mathbb{E}[\text{trace}(\vct{x}_T^T(\mtx{W}_T-\widehat{\mtx{W}_T})^T(\mtx{W}_T-\widehat{\mtx{W}_T})\vct{x}_T)]+k\sigma^2\nonumber\\
    &=\mathbb{E}[\text{trace}((\mtx{W}_T-\widehat{\mtx{W}_T})^T(\mtx{W}_T-\widehat{\mtx{W}_T})\vct{x}_T\vct{x}_T^T)]+k\sigma^2\nonumber\\
    &=\text{trace}((\mtx{W}_T-\widehat{\mtx{W}_T})^T(\mtx{W}_T-\widehat{\mtx{W}_T})\mathbb{E}[\vct{x}_T\vct{x}_T^T])+k\sigma^2\nonumber\\
    &=\text{trace}((\mtx{W}_T-\widehat{\mtx{W}_T})^T(\mtx{W}_T-\widehat{\mtx{W}_T})\mtx{\Sigma}_T)+k\sigma^2\nonumber\\
    &=||\mtx{\Sigma}_T^{\frac{1}{2}}(\mtx{W}_T-\widehat{\mtx{W}_T})^T||_F^2+k\sigma^2
\end{align}

\begin{itemize}[leftmargin=*]
    \item \textbf{One-hidden layer neural network model with fixed hidden-to-output layer:}
\end{itemize}
By expanding the expression we obtain

\begin{align}\label{gen1}
\mathbb{E}_{\mathbb{Q}_{\mtx{\theta}_T}}[\twonorm{\widehat{\vct{y}}_T-\vct{y}_T}^2]&=\mathbb{E}[\twonorm{\mtx{V}\varphi(\widehat{\mtx{W}}_T\vct{x}_T)-\mtx{V}\varphi(\mtx{W}_T\vct{x}_T)}^2]+k\sigma^2\nonumber\\
&\geq \sigma^2_{\text{min}}(\mtx{V})\mathbb{E}[\twonorm{\varphi(\widehat{\mtx{W}}_T\vct{x}_T)-\varphi(\mtx{W}_T\vct{x}_T)}^2]+k\sigma^2
\end{align}

Let $\mtx{A}=\widehat{\mtx{W}}_T\mtx{\Sigma}^{\frac{1}{2}}_T, \mtx{B}=\mtx{W}_T\mtx{\Sigma}^{\frac{1}{2}}_T$, and $\vct{x}=\mtx{\Sigma}^{\frac{-1}{2}}_T\vct{x}_T$. So $\vct{x}\sim \mathcal{N}(0,I_d)$. Moreover, let $\mtx{A}=
 \left [\begin{array}{cc}
   \vct{\alpha}_1^T  \\
   \vdots \\
   \vct{\alpha}_{\ell}^T  \\
  \end{array}\right]$ and $\mtx{B}=
 \left [\begin{array}{cc}
   \vct{\beta}_1^T  \\
   \vdots \\
   \vct{\beta}_{\ell}^T  \\
  \end{array}\right]$. Since $\mathbb{E}[\twonorm{\varphi(\widehat{\mtx{W}}_T\vct{x}_T)-\varphi(\mtx{W}_T\vct{x}_T)}^2]=\sum_{i=1}^{\ell}\mathbb{E}[|\varphi(\vct{\alpha}_i^T\vct{x})-\varphi(\vct{\beta}_i^T\vct{x})|^2]$, it suffices to find a lower bound for the following expression
  
  $$\mathbb{E}[|\varphi(\vct{a}^T\vct{x})-\varphi(\vct{b}^T\vct{x})|^2]$$
  
  where $a$ and $b$ are two arbitrary vectors in $\mathbb{R}^d$, $\varphi$ is the $\text{ReLU}$ activation function, and $\vct{x}\sim \mathcal{N}(0,I_d)$.
  
  We have
  \begin{align}\label{exprelu}
      \mathbb{E}[|\varphi(\vct{a}^T\vct{x})-\varphi(\vct{b}^T\vct{x})|^2]&=\mathbb{E}[|\varphi(\vct{a}^T\vct{x})|^2]+ \mathbb{E}[|\varphi(\vct{b}^T\vct{x})|^2]-2\mathbb{E}[\varphi(\vct{a}^T\vct{x})\varphi(\vct{b}^T\vct{x})].    
  \end{align}
  Now we calculate each term appearing on the right hand side.
  
  Since $\vct{a}^T\vct{x}\sim N(0,\twonorm{\vct{a}}^2)$, we have
  
  \begin{align*}
  \mathbb{E} [|\varphi(\vct{a}^T\vct{x})|^2]&=\mathbb{E}[|\text{ReLU}(\vct{a}^T\vct{x})|^2]\\
  &=\int_{0}^{+\infty}\frac{t^2}{\sqrt{2\pi}\twonorm{\vct{a}}}e^{\frac{-t^2}{2\twonorm{\vct{a}}^2}}dt\\
  &=\frac{\twonorm{\vct{a}}^2}{2}.
\end{align*}
Similarly, $\mathbb{E} [|\varphi(\vct{b}^T\vct{x})|^2]=\frac{\twonorm{\vct{b}}^2}{2}$.
To calculate the cross term note that $\vct{a}^T\vct{x}$ and $\vct{b}^T\vct{x}$ are jointly Gaussian with zero mean and covariance matrix equal to
\[
  \left[ {\begin{array}{cc}
   \twonorm{\vct{a}}^2 & \vct{a}^T\vct{b} \\
   \vct{a}^T\vct{b} & \twonorm{\vct{b}}^2 \\
  \end{array} } \right].
\]
Therefore, we have (e.g.~see \cite{daniely2016toward})
\begin{align}\label{relu-exp}
    2\mathbb{E}[\varphi(\vct{a}^T\vct{x})\varphi(\vct{b}^T\vct{x})]&=2\mathbb{E}[\text{ReLU}(\vct{a}^T\vct{x})\text{ReLU}(\vct{b}^T\vct{x})]\nonumber\\
    &=\twonorm{\vct{a}}\twonorm{\vct{b}}\frac{\sqrt{1-\gamma^2}+(\pi-\cos^{-1}(\gamma))\gamma}{\pi}
\end{align}
where $\gamma:=\frac{\vct{a}^T\vct{b}}{\twonorm{\vct{a}}\twonorm{\vct{b}}}$.
\\
Plugging these results in \eqref{exprelu}, we can conlude that 
\begin{align}\label{exprelu1}
   \mathbb{E}[|\varphi(\vct{a}^T\vct{x})-\varphi(\vct{b}^T\vct{x})|^2]&=\frac{\twonorm{\vct{a}}^2}{2}+\frac{\twonorm{\vct{b}}^2}{2}-\twonorm{\vct{a}}\twonorm{\vct{b}}\frac{\sqrt{1-\gamma^2}+(\pi-\cos^{-1}(\gamma))\gamma}{\pi}\nonumber\\
   &=\frac{1}{2}\twonorm{\vct{a}-\vct{b}}^2-\twonorm{\vct{a}}\twonorm{\vct{b}}\frac{\sqrt{1-\gamma^2}-\gamma\cos^{-1}(\gamma)}{\pi}.
\end{align}
We are interested in finding a universal constant $0<c<\frac{1}{2}$ such that $\mathbb{E}[|\varphi(\vct{a}^T\vct{x})-\varphi(\vct{b}^T\vct{x})|^2]\geq c\twonorm{\vct{a}-\vct{b}}^2$. Using \eqref{exprelu1} and dividing by $\twonorm{\vct{a}}\twonorm{\vct{b}}$ this is equivalent to finding $0<c<\frac{1}{2}$ such that
\begin{align*}
    \left(\frac{1}{2}-c\right)\frac{\twonorm{\vct{a}}^2+\twonorm{\vct{b}}^2-2\vct{a}^T\vct{b}}{\twonorm{\vct{a}}\twonorm{\vct{b}}}+\frac{\gamma\cos^{-1}(\gamma)-\sqrt{1-\gamma^2}}{\pi}\geq 0
\end{align*}
Next note that by the AM-GM inequality we have
\begin{align*}
   (\frac{1}{2}-c)\frac{\twonorm{\vct{a}}^2+\twonorm{\vct{b}}^2-2\vct{a}^T\vct{b}}{\twonorm{\vct{a}}\twonorm{\vct{b}}}+&\frac{\gamma\cos^{-1}(\gamma)-\sqrt{1-\gamma^2}}{\pi}\\
   &\geq (\frac{1}{2}-c)\frac{2\twonorm{\vct{a}}\twonorm{\vct{b}}-2\vct{a}^T\vct{b}}{\twonorm{\vct{a}}\twonorm{\vct{b}}}+\frac{\gamma\cos^{-1}(\gamma)-\sqrt{1-\gamma^2}}{\pi}\\
   &=(\frac{1}{2}-c)(2-2\gamma)+\frac{\gamma\cos^{-1}(\gamma)-\sqrt{1-\gamma^2}}{\pi}\\
   &=(1-\gamma)\big[(1-2c)+\frac{1}{\pi}\cdot\frac{\gamma\cos^{-1}(\gamma)-\sqrt{1-\gamma^2}}{1-\gamma}\big].
\end{align*}
Therefore, it suffices to find $0<c<\frac{1}{2}$ such that the R.H.S. of the above is positive.
%Note that if $\gamma=1$, in \eqref{exprelu1} we get $\mathbb{E}[|\varphi(\vct{a}^T\vct{x})-\varphi(\vct{b}^T\vct{x})|^2]=\frac{1}{2}\twonorm{\vct{a}-\vct{b}}^2$.
It is easy to verify that $h(\gamma):=\frac{\gamma\cos^{-1}(\gamma)-\sqrt{1-\gamma^2}}{1-\gamma}\geq \frac{-\pi}{2}$ for $-1\leq\gamma<1$. This in turn implies that the R.H.S. above is positive with $c=\frac{1}{4}$.

In the case when $\twonorm{\vct{a}}=0$ or $\twonorm{\vct{b}}=0$ ( let us assume $\twonorm{\vct{b}}=0$),  \eqref{exprelu} reduces to 

\begin{align*}
      \mathbb{E}[|\varphi(\vct{a}^T\vct{x})-\varphi(\vct{b}^T\vct{x})|^2]&=\mathbb{E}[|\varphi(\vct{a}^T\vct{x})|^2]\nonumber\\
      &=\frac{\twonorm{\vct{a}}^2}{2}\nonumber\\
      &\geq \frac{1}{2}\twonorm{\vct{a}-\vct{b}}^2\\
      &\geq \frac{1}{4}\twonorm{\vct{a}-\vct{b}}^2.
  \end{align*}
Plugging the latter into \eqref{gen1} we arrive at 
\begin{align*}
\mathbb{E}_{\mathbb{Q}_{\theta_T}}[\twonorm{\widehat{\vct{y}}_T-\vct{y}_T}^2]&\geq \sigma^2_{\text{min}}(\mtx{V})\mathbb{E}[\twonorm{\varphi(\widehat{\mtx{W}}_T\vct{x}_T)-\varphi(\mtx{W}_T\vct{x}_T)}^2]+k\sigma^2\\
&\geq \frac{1}{4}\sigma^2_{\text{min}}(\mtx{V})||\mtx{\Sigma}_T^{\frac{1}{2}}(\widehat{\mtx{W}}_T-\mtx{W}_T)^T||_F^2+k\sigma^2,
\end{align*}
concluding the proof.

\begin{itemize}[leftmargin=*]
    \item \textbf{One-hidden layer neural network model with fixed input-to-hidden layer:}
\end{itemize}
By expanding the expression we get 
\begin{align*}
\mathbb{E}_{\mathbb{Q}_{\mtx{\theta}_T}}[\twonorm{\widehat{\vct{y}}_T-\vct{y}_T}^2]&=\mathbb{E}[\twonorm{\widehat{\mtx{V}}_T\varphi(\mtx{W}\vct{x}_T)-\mtx{V}_T\varphi(\mtx{W}\vct{x}_T)}^2]+k\sigma^2.
\end{align*}
If we denote $\mathbb{E}[\varphi(\mtx{W}\vct{x}_T)\varphi(\mtx{W}\vct{x}_T)^T]=\widetilde{\mtx{\Sigma}}_T$, then similar to \eqref{linear-gen} we obtain
\begin{align}
    \mathbb{E}_{\mathbb{Q}_{\mtx{\theta}_T}}[\twonorm{\widehat{\vct{y}}_T-\vct{y}_T}^2]&=||\widetilde{\mtx{\Sigma}}_T^{\frac{1}{2}}(\widehat{\mtx{V}}_T-\mtx{V}_T)^T||_F^2+k\sigma^2.
\end{align}
Therefore, it suffices to calculate $\widetilde{\mtx{\Sigma}}_T$. Let $\mtx{W}\mtx{\Sigma}^{\frac{1}{2}}_T=
 \left [\begin{array}{cc}
   \vct{a}_1^T  \\
   \vdots \\
   \vct{a}_{\ell}^T  \\
  \end{array}\right]$ and $\vct{x}=\mtx{\Sigma}^{\frac{-1}{2}}_T\vct{x}_T$ (so $\vct{x}\sim \mathcal{N}(0,I_d)$). By \eqref{relu-exp} we obtain that
  \begin{align}\label{covariance}
\widetilde{\mtx{\Sigma}}_T=\big[\frac{1}{2}\twonorm{\vct{a}_i}\twonorm{\vct{a}_j}\frac{\sqrt{1-\gamma_{ij}^2}+(\pi-\cos^{-1}(\gamma_{ij}))\gamma_{ij}}{\pi}\big]_{ij}
\end{align}
  where $\gamma_{ij}:=\frac{\vct{a}_i^T\vct{a}_j}{\twonorm{\vct{a}_i}\twonorm{\vct{a}_j}}$.

\subsection{Calculating KL-Divergences for the Linear Model (Proof of Lemma \ref{KLdiv})}\label{prooflemma1}
First we compute the KL-divergence between the distributions $\mathbb{P}_{\mtx{W}_S^{(i)}}(\vct{x}_S,\vct{y}_S)$ and $\mathbb{P}_{\mtx{W}_S^{(j)}}(\vct{x}_S,\vct{y}_S)$:
\begin{align*}
D_{KL}(\mathbb{P}_{\mtx{W}_S^{(i)}}(\vct{x}_S,\vct{y}_S),\mathbb{P}_{\mtx{W}_S^{(j)}}(\vct{x}_S,\vct{y}_S))=&D_{KL}(\mathbb{P}_{\mtx{W}_S^{(i)}}(\vct{x}_S),\mathbb{P}_{\mtx{W}_S^{(j)}}(\vct{x}_S))\\
&+\mathbb{E}[D_{KL}(\mathbb{P}_{\mtx{W}_S^{(i)}}(\vct{y}_S|\vct{x}_S),\mathbb{P}_{\mtx{W}_S^{(j)}}(\vct{y}_S|\vct{x}_S))].
\end{align*}

The marginal distributions $\mathbb{P}_{\mtx{W}_S^{(i)}}(\vct{x}_S)$ and $\mathbb{P}_{\mtx{W}_S^{(j)}}(\vct{x}_S)$ are equal so their KL-divergence is zero. The conditional distributions $\mathbb{P}_{\mtx{W}_{S}^{(i)}}(\vct{y}_{S}|\vct{x}_{S})$ and $\mathbb{P}_{\mtx{W}_{S}^{(j)}}(\vct{y}_{S}|\vct{x}_{S})$ 
are normally distributed with covariance matrix $\sigma^2\mathbf{I}_k$ and with mean respectively equal to $\mtx{W}_S^{(i)} \vct{x}_S$ and $\mtx{W}_S^{(j)} \vct{x}_S$. Therefore, 
$$D_{KL}(\mathbb{P}_{\mtx{W}_S^{(i)}}(\vct{y}_S|\vct{x}_S),\mathbb{P}_{\mtx{W}_S^{(j)}}(\vct{y}_S|\vct{x}_S))=\frac{\twonorm{\mtx{W}_S^{(i)}\vct{x}_S-\mtx{W}_S^{(j)}\vct{x}_S}^2}{2\sigma^2}.$$
This in turn implies that 
$$D_{KL}(\mathbb{P}_{\mtx{W}_S^{(i)}}(\vct{x}_S,\vct{y}_S),\mathbb{P}_{\mtx{W}_S^{(j)}}(\vct{x}_S,\vct{y}_S))=\frac{\mathbb{E}[\twonorm{\mtx{W}_S^{(i)}\vct{x}_S-\mtx{W}_S^{(j)}\vct{x}_S}^2]}{2\sigma^2}=\frac{||\mtx{\Sigma}_S^{\frac{1}{2}}(\mtx{W}_S^{(i)}-\mtx{W}_S^{(j)})^T||_F^2}{2\sigma^2},$$
where the last equality follows similarly to the proof of Proposition \ref{prop1} in the linear case.

A similar calculation also yields
$$D_{KL}(\mathbb{Q}_{\mtx{W}_T^{(i)}}(\vct{x}_T,\vct{y}_T),\mathbb{Q}_{\mtx{W}_T^{(j)}}(\vct{x}_T,\vct{y}_T))=\frac{||\mtx{\Sigma}_T^{\frac{1}{2}}(\mtx{W}_T^{(i)}-\mtx{W}_T^{(j)})^T||_F^2}{2\sigma^2}.$$
\subsection{Lower Bound for Minimax Risk When $\Delta\geq \sqrt{\frac{\sigma^2D\log{2}}{r_{T}n_{T}}}$ and $\Delta<\sqrt{\frac{\sigma^2D\log{2}}{r_{T}n_T}}$ ( Proof of Lemma \ref{b1})}\label{prooflemma2}
Consider the set
\begin{align*}
\Big\{\eta : \eta=\mtx{\Sigma}_T^{\frac{1}{2}}\mtx{W}_T^T \text{for some}\ \mathbf{W}_T \in \mathbb{R}^{k \times d} \ \text{and} \ ||\eta||_F\leq 4\delta\Big\}
\end{align*}
where $\delta>0$ is a value to be determined later in the proof. Furthermore, let $\{\eta^1,...,\eta^N\}$ be a $2\delta$-packing of this set in the $F$-norm. Since dim(range$(\mtx{\Sigma}_T^{\frac{1}{2}}\mtx{W}_T^T)$)$=rk$ in which $\mtx{W}_T$ is regarded as an input, this set sits in a space of dimension $rk$, where $r=$rank($\mtx{\Sigma}_T$). Hence we can find such a packing with $\log N\geq rk \log{2}$ elements.

Therefore, we have a collection of matrices of the form $\eta^j=\mtx{\Sigma}_T^{\frac{1}{2}}(\mtx{W}_T^{(i)})^T$ for some $\mtx{W}_T^{(i)} \in \mathbb{R}^{k \times d}$ such that
\begin{align*}
\fronorm{\mtx{\Sigma}_T^{\frac{1}{2}}(\mtx{W}_T^{(i)})^T}\leq 4\delta\ \text{for each}\ i\in [N]
\end{align*}
and
\begin{align*}
2\delta\leq ||\mtx{\Sigma}_T^{\frac{1}{2}}(\mtx{W}_T^{(i)}-\mtx{W}_T^{(j)})^T||_F\leq 8\delta \ \text{for each}\ i\neq j \in [N]\times [N].    
\end{align*}
So by Lemma \ref{KLdiv} we get
$$D_{KL}(\mathbb{Q}_{\mtx{W}_T^{(i)}},\mathbb{Q}_{\mtx{W}_T^{(j)}})\leq \frac{32\delta^2}{\sigma^2} \ \text{for each}\ i\neq j \in [N]\times [N].$$
Then set $\delta\leq \frac{\Delta}{8}$. We can choose $\mathbb{P}_{\mtx{W}_S^{(1)}}=...=\mathbb{P}_{\mtx{W}_S^{(N)}}$ with $\mtx{W}_S^{(1)}=...=\mtx{W}_S^{(N)}$ and
$||\mtx{\Sigma}_T^{\frac{1}{2}}(\mtx{W}_S^{(1)})^T||_F= 4\delta$
. So they satisfy the condition 
\begin{align*}
    \rho(\mtx{W}_S^{(i)},\mtx{W}_T^{(i)})\leq 8\delta \leq \Delta \ \  \text{for each}\ i\in [N].
\end{align*}
\begin{figure}[t]
\centering

\includegraphics[scale=0.56]{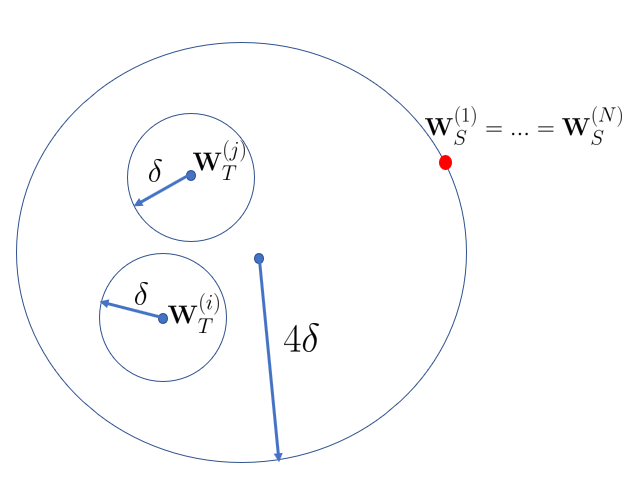}

\caption{Configuration of the parameters of source and target distributions in Lemma \ref{b1}.}
\label{fig4}

\end{figure}
Figure \ref{fig4} illustrates this configuration. So having samples from $\mathbb{P}_{\mtx{W}_S^{(i)}}$ does not contain any information about the true index in $[N]$ in the hypothesis testing problem and $E$ is independent of $J$, hence we get
$$I(J;E)=0.$$
Therefore, using \eqref{minmax1} and \eqref{info} we get 
\begin{align*}
    \mathcal{R}_T(\mathcal{P}_{\Delta};\phi \circ \rho)&\geq \phi(\delta)\left(1-\frac{n_{T}\frac{32\delta^2}{\sigma^2}+\log{2}}{rk\log{2}}\right)\nonumber\\
    &=\delta^2\left(1-\frac{n_{T}\frac{32\delta^2}{\sigma^2}+\log{2}}{rk\log{2}}\right)
\end{align*}
for any $0\leq \delta\leq \frac{\Delta}{8}$. We need to check the boundary and stationary points to solve the above optimization problem.

If $\Delta\geq \sqrt{\frac{\sigma^2(rk-1)\log{2}}{n_{T}}}=\sqrt{\frac{\sigma^2D\log{2}}{n_{T}}}$ holds, then
\begin{align*}
   \mathcal{R}_T(\mathcal{P}_{\Delta};\phi \circ \rho)\geq \frac{\sigma^2 (rk-1)^2\log{2}}{128n_{T}rk}.
\end{align*}
Since $D=rl-1\geq 20$, we have $\frac{\log{2}}{rk}\geq \frac{1}{2(rk-1)}$, so
\begin{align}
     \mathcal{R}_T(\mathcal{P}_{\Delta};\phi \circ \rho)\geq \frac{\sigma^2D}{256n_{\mathbb{Q}}}
\end{align}
and if $\Delta<\sqrt{\frac{\sigma^2(rk-1)\log{2}}{n_T}}=\sqrt{\frac{\sigma^2D\log{2}}{n_{T}}}$ then
\begin{align*}
   \mathcal{R}_T(\mathcal{P}_{\Delta};\phi \circ \rho)&\geq (\frac{\Delta}{8})^2[1-\frac{\frac{n_{T}\Delta^2}{2\sigma^2}+\log{2}}{rk\log{2}}]\\
   &\geq(\frac{\Delta}{8})^2[1-\frac{\frac{n_{T}\Delta^2}{2\sigma^2}+\log{2}}{D\log{2}}].
\end{align*}
Since $D\geq 20$ we get
\begin{align}
    \mathcal{R}_T(\mathcal{P}_{\Delta};\phi \circ \rho)\geq \frac{1}{100}\Delta^2[1-0.8\frac{n_{T}\Delta^2}{\sigma^2D}].
\end{align}

\subsection{Lower Bound for Minimax Risk When $\Delta\leq \frac{1}{45}\sqrt{\frac{\sigma^2D}{r_{S}n_{S}+r_{T}n_{T}}}$ ( Proof of Lemma \ref{b2})}\label{prooflemma3}
Let $\delta'=\Delta+\underbrace{u\Delta}_{=\delta}$, for $u>0$ to be determined, Consider the set
$$\{\eta : \eta=\mtx{\Sigma}_T^{\frac{1}{2}}\mtx{W}_S^T \ \text{for some}\ \mathbf{W}_S \in \mathbb{R}^{k \times d} \ \text{and} \ ||\eta||_F\leq 4\delta'\}$$
and let $\{\eta^1,...,\eta^N\}$ be a $2\delta'$-packing in the $F$-norm and consider each $\eta^i$ as a single point. Since dim(range$(\mtx{\Sigma}_T^{\frac{1}{2}}\mtx{W}_T^T)$)$=rk$ in which $\mtx{W}_T$ is regarded as an input, this set sits in a space of dimension $rk$ where $r=$rank($\Sigma_T$). Therefore, we can find such a packing with $\log N\geq rk \log{2}$ elements.

Hence, we have a collection of matrices of the form $\eta^i=\mtx{\Sigma}_T^{\frac{1}{2}}(\mtx{W}_S^{(i)})^T$ for some $\mtx{W}_S^{(i)} \in \mathbb{R}^{k \times d}$ such that
$$||\mtx{\Sigma}_T^{\frac{1}{2}}(\mtx{W}_S^{(i)})^T||_F\leq 4\delta'\ \text{for each}\ i\in [N]$$

$$2\delta'\leq ||\mtx{\Sigma}_T^{\frac{1}{2}}(\mtx{W}_S^{(i)}-\mtx{W}_S^{(j)})^T||_F\leq 8\delta' \ \text{for each}\ i\neq j \in [N]\times [N].$$

\begin{figure}[htp]
\centering

\includegraphics[scale=0.63]{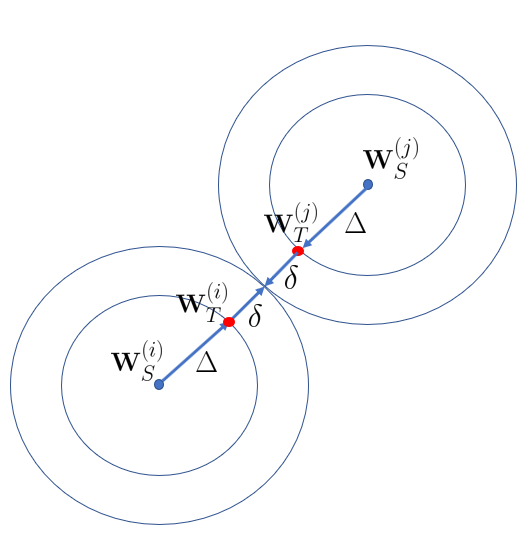}

\caption{Configuration of the parameters of source and target distributions in Lemma \ref{b2}.}
\label{fig5}

\end{figure}

We choose each $\mtx{W}_T^{(i)}$ such that $\rho(\mtx{W}_T^{(i)},\mtx{M}_S^{(i)})=||\mtx{\Sigma}_T^{\frac{1}{2}}(\mtx{W}_T^{(i)}-\mtx{W}_S^{(i)})^T||_F=\Delta$. So

$$\rho(\mtx{W}_T^{(i)},\mtx{W}_T^{(j)})\geq 2\delta \ \text{for each}\ i\neq j \in [N]\times [N].$$
Moreover, 
\begin{align*}
    \rho(\mtx{W}_T^{(i)},\mtx{W}_T^{(j)})&\leq \rho(\mtx{W}_T^{(i)},\mtx{W}_S^{(i)})+\rho(\mtx{W}_S^{(i)},\mtx{M}_S^{(j)})+\rho(\mtx{W}_S^{(j)},\mtx{W}_T^{(j)})\\
    &\leq 2\Delta+8(\Delta+u\Delta).
\end{align*}
Figure \ref{fig5} illustrates this configuration. By Lemma \ref{KLdiv} we have
$$D_{KL}(\mathbb{Q}_{\mtx{W}_T^{(i)}},\mathbb{Q}_{\mtx{W}_T^{(j)}})\leq \frac{2\Delta^2(5+4u)^2}{\sigma^2}  \ \text{for each}\ i\neq j \in [N]\times [N].$$

Also we have

\begin{align*}
    ||\mtx{\Sigma}_S^{\frac{1}{2}}(\mtx{W}_S^{(i)}-\mtx{W}_S^{(j)})^T||_F&=||\mtx{\Sigma}_S^{\frac{1}{2}}\mtx{\Sigma}_T^{-\frac{1}{2}}\mtx{\Sigma}_T^{\frac{1}{2}}(\mtx{W}_S^{(i)}-\mtx{W}_S^{(j)})^T||_F\\
    &\leq 8\opnorm{\mtx{\Sigma}_S^{\frac{1}{2}}\mtx{\Sigma}_T^{-\frac{1}{2}}}\delta'\\
    &=8\opnorm{\mtx{\Sigma}_S^{\frac{1}{2}}\mtx{\Sigma}_T^{-\frac{1}{2}}}(\Delta+u\Delta)
\end{align*}

Hence, by Lemma \ref{KLdiv} we have
$$D_{KL}(\mathbb{P}_{\mtx{W}_S^{(i)}},\mathbb{P}_{\mtx{W}_S^{(j)}})\leq \frac{32\opnorm{\mtx{\Sigma}_S^{\frac{1}{2}}\mtx{\Sigma}_T^{-\frac{1}{2}}}^2\Delta^2(u+1)^2}{\sigma^2} \ \text{for each}\ i\neq j \in [N]\times [N].$$

Therefore, using \eqref{minmax1} and \eqref{info} we arrive at 
\begin{align*}
    \mathcal{R}_T(\mathcal{P};\phi \circ \rho)&\geq \phi(u\Delta)[1-\frac{n_{S}\frac{32\opnorm{\Sigma_S^{\frac{1}{2}}\Sigma_T^{-\frac{1}{2}}}^2\Delta^2(u+1)^2}{\sigma^2}+n_{T}\frac{2\Delta^2(5+4u)^2}{\sigma^2}+\log{2}}{rk\log{2}}]\nonumber\\
    &=(u\Delta)^2[1-\frac{n_{S}r_{S}\frac{32\Delta^2(u+1)^2}{\sigma^2}+n_{T}r_T\frac{2\Delta^2(5+4u)^2}{\sigma^2}+\log{2}}{rk\log{2}}].
\end{align*}
The above inequality holds for every $u\geq 0$. Maximizing the expression above over $u$ we can conclude 
if $\Delta\leq \sqrt{\frac{\sigma^2(rk-1)\log{2}}{32n_{S}r_{S}+50n_{T}r_T}}$ then
\begin{align*}
   \mathcal{R}_T(\mathcal{P};\phi \circ \rho)\geq (u\Delta)^2[1-\frac{n_{S}r_{S}\frac{32\Delta^2(u+1)^2}{\sigma^2}+n_{T}r_T\frac{2\Delta^2(5+4u)^2}{\sigma^2}+\log{2}}{rk\log{2}}]
\end{align*}
where $u=\frac{3\Delta(4n_{S}r_{S}+n_{T}r_T)+\sqrt{\Delta^2[16(n_{S}r_{S})^2+25(n_{T}r_T)^2+32n_{S}r_{S}n_{T}r_T]+4(n_{S}r_{S}+n_{T}r_T)(rk-1)\sigma^2\log{2}}}{16\Delta(n_{\mathbb{P}}r_{\mathbb{P}}+n_{\mathbb{Q}})}.$

Now, we need to simplify the above expressions. First note that 
\begin{align*}
    (u\Delta)&\geq (\frac{3\Delta}{16}+\frac{\sqrt{\Delta^2+4\frac{D\sigma^2\log2}{n_{S}r_{S}+n_{T}r_T}}}{16}),
\end{align*}
so
\begin{align}
    (u\Delta)^2\geq \frac{\Delta^2+2.7\frac{D\sigma^2}{n_{S}r_{S}+n_{T}r_T}}{256}.
\end{align}
Moreover, 
\begin{align*}
    1-\frac{n_{S}r_{S}\frac{32\Delta^2(u+1)^2}{\sigma^2}+n_{T}\frac{2\Delta^2(5+4u)^2}{\sigma^2}+\log{2}}{rk\log{2}}&\geq  1-\frac{[n_{S}r_{S}+n_{T}r_T]\frac{32\Delta^2(\frac{5}{4}+u)^2}{\sigma^2}+\log{2}}{D\log{2}}
\end{align*}
and 
\begin{align*}
    \Delta(\frac{5}{4}+u)\leq 2\Delta+\frac{1}{16}\sqrt{25\Delta^2+\frac{4\log{2}D\sigma^2}{n_{S}r_{S}+n_{T}r_T}}.
\end{align*}
Since $\Delta\leq \frac{1}{45}\sqrt{\frac{\sigma^2D}{n_{S}r_{S}+n_{T}r_T}}$,
\begin{align*}
    \Delta^2(\frac{5}{4}+u)^2&\leq (4+(\frac{5}{16})^2)\Delta^2+\frac{1}{4}\Delta\sqrt{25\Delta^2+\frac{4\log{2}D\sigma^2}{n_{S}r_{S}+n_{T}r_T}}\\
    &\leq (4+(\frac{5}{16})^2)\frac{1}{45^2}\frac{D\sigma^2}{n_{S}r_{S}+n_{T}r_T}+\frac{1}{4\times 45^2}\sqrt{25^2+45^2\times 4\log{2}}\frac{D\sigma^2}{n_{S}r_{S}+n_{T}r_T}\\
    &\leq 0.012\frac{D\sigma^2}{n_{S}r_{S}+n_{T}r_T}.
\end{align*}
Hence, 
\begin{align*}
    1-\frac{[n_{S}r_{S}+n_{T}r_T]\frac{32\Delta^2(\frac{5}{4}+u)^2}{\sigma^2}+\log{2}}{D\log{2}}&\geq 1-0.56-\frac{1}{D}\\
    &\geq 0.39.
\end{align*}
Therefore, we arrive at
\begin{align}
    \mathcal{R}_T(\mathcal{P};\phi \circ \rho)\geq \frac{\Delta^2}{1000}+\frac{6}{1000}\frac{D\sigma^2}{n_{S}r_{S}+n_{T}r_T}.
\end{align}

\subsection{Proof of Theorem \ref{thm1} (One-hidden layer neural network with fixed hidden-to-output layer)}\label{proofmodel2}
By Proposition \ref{prop1}, the generalization error is bounded from below as 
\begin{align*}
\mathbb{E}_{\mathbb{Q}_{\vct{\theta}_T}}[\twonorm{\widehat{\vct{y}}_T-\vct{y}_T}^2]\geq \frac{1}{4}\sigma^2_{\text{min}}(\mtx{V})||\mtx{\Sigma}_T^{\frac{1}{2}}(\widehat{\mtx{W}}_T-\mtx{W}_T)^T||_F^2+k\sigma^2.
\end{align*}
Therefore, it suffices to find a lower bound for the following quantity:
\begin{align*}
\mathcal{R}_T(\mathcal{P}_{\Delta}&;\phi \circ \rho)\\
&:=\inf_{\widehat{\mtx{W}}_T}\sup_{(\mathbb{P}_{\mtx{W}_S}, \mathbb{Q}_{\mtx{W}_T}) \in \mathcal{P}_{\Delta}}\mathbb{E}_{S_{\mathbb{P}_{\mtx{W}_S}}\sim \mathbb{P}_{\mtx{W}_S}^{1:n_{\mathbb{P}}}}\big[\mathbb{E}_{S_{\mathbb{Q}_{\mtx{W}_T}}\sim \mathbb{Q}_{\mtx{W}_T}^{1:n_{\mathbb{Q}}}}\big[\phi(\rho(\widehat{\mtx{W}}_T(S_{\mathbb{P}_{\mtx{W}_S}},S_{\mathbb{Q}_{\mtx{W_T}}}),\mtx{W_T}))\big]\big]
\end{align*}
where $\phi(x)=x^2$ for $x\in \mathbb{R}$ and $\rho$ is defined per Definition \ref{definition1}.
The rest of the proof is similar to the linear case as the corresponding transfer distance metrics are the same. We only need to upper bound the corresponding KL-divergences in this case. We do so by the following lemma.

\begin{lemma}\label{KLdiv2}
Suppose that $\mathbb{P}_{\mtx{W}_S^{(i)}}$ and $\mathbb{P}_{\mtx{W}_S^{(j)}}$ are the joint distributions of features and labels in a source task and $\mathbb{Q}_{\mtx{W}_T^{(i)}}$ and $\mathbb{Q}_{\mtx{W}_T^{(j)}}$ are joint distributions of features and labels in a target task as defined in Section \ref{sec2.1} in the one-hidden layer neural network with fixed hidden-to-output layer model. Then  $D_{KL}(\mathbb{P}_{\mtx{W}_S^{(i)}}||\mathbb{P}_{\mtx{W}_S^{(j)}})\leq\frac{\opnorm{\mtx{V}}^2||\mtx{\Sigma}_S^{\frac{1}{2}}(\mtx{W}_S^{(i)}-\mtx{W}_S^{(j)})^T||_F^2}{2\sigma^2}$ and $D_{KL}(\mathbb{Q}_{\mtx{W}_T^{(i)}}||\mathbb{Q}_{\mtx{W}_T^{(j)}})\leq\frac{\opnorm{\mtx{V}}^2||\mtx{\Sigma}_S^{\frac{1}{2}}(\mtx{W}_T^{(i)}-\mtx{W}_T^{(j)})^T||_F^2}{2\sigma^2}.$

\end{lemma}

Furthermore, we also note that since in this case $\mtx{W}_S, \mtx{W}_T\in \mathbb{R}^{\ell \times d}$, the definition of $D$ is slightly different from that in the linear case. In this case $D=\text{rank}(\mtx{\Sigma}_T)\ell-1$.

\subsection{Bounding the KL-Divergences in the Neural Network Model (Proof of Lemma \ref{KLdiv2})}
First we compute the KL-divergence between the distributions $\mathbb{P}_{\mtx{W}_S^{(i)}}(\vct{x}_S,\vct{y}_S)$ and $\mathbb{P}_{\mtx{W}_S^{(j)}}(\vct{x}_S,\vct{y}_S)$
\begin{align*}
D_{KL}(\mathbb{P}_{\mtx{W}_S^{(i)}}(\vct{x}_S,\vct{y}_S),\mathbb{P}_{\mtx{W}_S^{(j)}}(\vct{x}_S,\vct{y}_S))=&D_{KL}(\mathbb{P}_{\mtx{W}_S^{(i)}}(\vct{x}_S),\mathbb{P}_{\mtx{W}_S^{(j)}}(\vct{x}_S))\\
&+\mathbb{E}[D_{KL}(\mathbb{P}_{\mtx{W}_S^{(i)}}(\vct{y}_S|\vct{x}_S),\mathbb{P}_{\mtx{W}_S^{(j)}}(\vct{y}_S|\vct{x}_S))].
\end{align*}

The marginal distributions $\mathbb{P}_{\mtx{M}_S^{(i)}}(\vct{x}_S)$ and $\mathbb{P}_{\mtx{M}_S^{(j)}}(\vct{x}_S)$ are equal so their KL-divergence is zero. The conditional distributions $\mathbb{P}_{\mtx{M}_S^{(i)}}(\vct{y}_S|\vct{x}_S)$ and $\mathbb{P}_{\mtx{M}_S^{(j)}}(\vct{y}_S|\vct{x}_S)$ are normally distributed with covariance matrix $\sigma^2\mathbf{I}_k$ and with mean respectively equal to $\mtx{V}\varphi(\mtx{W}_S^{(i)} \vct{x}_S)$ and $\mtx{V}\varphi(\mtx{W}_S^{(j)} \vct{x}_S)$. Therefore, we obtain  
$$D_{KL}(\mathbb{P}_{\mtx{W}_S^{(i)}}(\vct{y}_S|\vct{x}_S),\mathbb{P}_{\mtx{W}_S^{(j)}}(\vct{y}_S|\vct{x}_S))=\frac{\twonorm{\mtx{V}\varphi(\mtx{W}_S^{(i)}\vct{x}_S)-\mtx{V}\varphi(\mtx{W}_S^{(j)}\vct{x}_S)}^2}{2\sigma^2}.$$

Then we have 
\begin{align*}
D_{KL}(\mathbb{P}_{\mtx{W}_S^{(i)}}(\vct{x}_S,\vct{y}_S),\mathbb{P}_{\mtx{W}_S^{(j)}}(\vct{x}_S,\vct{y}_S))&=\frac{\mathbb{E}\twonorm{\mtx{V}\varphi(\mtx{W}_S^{(i)}\vct{x}_S)-\mtx{V}\varphi(\mtx{W}_S^{(j)}\vct{x}_S)}^2}{2\sigma^2}\\
&\leq\frac{\opnorm{\mtx{V}}^2||\mtx{\Sigma}_S^{\frac{1}{2}}(\mtx{W}_S^{(i)}-\mtx{W}_S^{(j)})^T||_F^2}{2\sigma^2}.
\end{align*}

Since ReLU is a Lipschitz function. Similarly we get 
\begin{align*}
D_{KL}(\mathbb{Q}_{\mtx{W}_T^{(i)}}(\vct{x}_T,\vct{y}_T),\mathbb{Q}_{\mtx{W}_T^{(j)}}(\vct{x}_T,\vct{y}_T))&=\frac{\mathbb{E}||\mtx{V}\varphi(\mtx{W}_T^{(i)}\vct{x}_T)-\mtx{V}\varphi(\mtx{W}_T^{(j)}\vct{x}_T)||_F^2}{2\sigma^2}\\
&\leq\frac{\opnorm{\mtx{V}}^2||\mtx{\Sigma}_T^{\frac{1}{2}}(\mtx{W}_T^{(i)}-\mtx{W}_T^{(j)})^T||_F^2}{2\sigma^2}.
\end{align*}

\subsection{Proof of Theorem \ref{thm1} (One-hidden layer neural network model with fixed input-to-hidden layer)}\label{proofmodel3}

By Proposition \ref{prop1}, the generalization error is 
$\mathbb{E}_{\mathbb{Q}_{\vct{\theta}_T}}[\twonorm{\widehat{\vct{y}}_T-\vct{y}_T}^2]=||\widetilde{\mtx{\Sigma}}_T^{\frac{1}{2}}(\widehat{\mtx{V}}_T-\mtx{V}_T)^T||_F^2+k\sigma^2$ so it suffices to find a lower bound for the following quantity 
\begin{align*}
\mathcal{R}_T(\mathcal{P}_{\Delta};\phi \circ \rho):=\inf_{\widehat{\mtx{V}}_T}\sup_{(\mathbb{P}_{\mtx{V}_S}, \mathbb{Q}_{\mtx{V}_T}) \in \mathcal{P}_{\Delta}}\mathbb{E}_{S_{\mathbb{P}_{\mtx{V}_S}}\sim \mathbb{P}_{\mtx{V}_S}^{1:n_{\mathbb{P}}}}\big[\mathbb{E}_{S_{\mathbb{Q}_{\mtx{V}_T}}\sim \mathbb{Q}_{\mtx{V}_T}^{1:n_{\mathbb{Q}}}}\big[\phi(\rho(\widehat{\mtx{V}}_T(S_{\mathbb{P}_{\mtx{V}_S}},S_{\mathbb{Q}_{\mtx{V}_T}}),\mtx{V}_T))\big]\big]
\end{align*}
Where $\phi(x)=x^2$ for $x\in \mathbb{R}$ and $\rho$ is defined per Definition \ref{definition1}. Inherently, this case is the same as the linear model except that the distribution of the features has changed which was calculated in \eqref{covariance}. The rest of the proof is similar to the linear case with the difference that $\mtx{\Sigma}_S, \mtx{\Sigma}_T$ should be replaced by $\widetilde{\mtx{\Sigma}}_S, \widetilde{\mtx{\Sigma}}_T$.

\end{document}